\documentclass{article}

\PassOptionsToPackage{numbers, compress}{natbib}

\newcommand{\link}[0]{%
    \ifdefined\anon
        \texttt{\url{https://github.com/sony/mucs}}%
    \else
        \texttt{\url{https://github.com/sony/mucs}}%
    \fi
}

\ifdefined\anon
    \usepackage[main, final]{neurips_2026}
\else
    \usepackage[preprint]{neurips_2026}
\fi

\usepackage[utf8]{inputenc} 
\usepackage[T1]{fontenc}    
\usepackage{hyperref}       
\usepackage{url}            
\usepackage{booktabs}       
\usepackage{amsfonts}       
\usepackage{nicefrac}       
\usepackage{microtype}      
\usepackage[dvipsnames]{xcolor}         

\usepackage{amsmath}
\usepackage{amssymb}
\usepackage{mathtools}
\usepackage{bm}

\newcommand{\se}[1]{\mathcal{#1}}

\newcommand{\ve}[1]{\textbf{#1}}
\newcommand{\tx}[1]{\text{#1}}
\newcommand{\mo}[1]{\mathop{\mathbb{#1}}}
\newcommand{\norm}[1]{\left\lVert#1\right\rVert}

\newcommand{\fb}[1]{\textbf{#1}}
\newcommand{\fu}[1]{\underline{#1}}

\definecolor{mydarkblue}{rgb}{0,0.08,0.45}
\hypersetup{ %
    pdftitle={},
    pdfkeywords={},
    pdfborder=0 0 0,
    pdfpagemode=UseNone,
    colorlinks=true,
    linkcolor=mydarkblue,
    citecolor=mydarkblue,
    filecolor=mydarkblue,
    urlcolor=mydarkblue,
}

\newcommand{\topic}[1]{{\noindent\textbf{#1 --- }}}

\usepackage{graphicx}
\usepackage{multirow}
\usepackage{colortbl}
\definecolor{gold}{RGB}{255,250,225}
\definecolor{silver}{RGB}{240,240,240}
\definecolor{bronze}{RGB}{250,240,235}
\definecolor{myorange}{RGB}{255,247,237}

\usepackage{fancyvrb}
\usepackage{varwidth}
\newcommand{\tcb}[1]{\textcolor{blue}{#1}}
\newcommand{\tcy}[1]{\textcolor{brown}{#1}}
\newcommand{\tcg}[1]{\textcolor{ForestGreen}{#1}}

\usepackage{placeins} 

\title{Training Data Attribution in Diffusion Models via Mirrored Unlearning and Noise-Consistent Skew}

\author{%
  Joan Serrà$^1$,\, Dipam Goswami$^1$\thanks{Work done during an internship at Sony AI.} ,\, Fabio Morreale$^1$,\, Wei-Hsiang Liao$^1$,\, and Yuki Mitsufuji$^{1,2}$ \\
  \\
  $^1$ Sony AI \\
  $^2$ Sony Group Corporation \\
  \texttt{name.surname@sony.com} \\
}

\begin{document}

\maketitle
\setcounter{footnote}{0}

\begin{abstract}
Training data attribution (TDA) should enable generative model interpretability and foster a variety of related downstream tasks. Nonetheless, current TDA approaches lack reliability and robustness, preventing their adoption in real-world setups. In this paper, we take a decisive step towards more reliable and robust TDA for diffusion models. We propose to perform TDA with mirrored unlearning and noise-consistent skew (MUCS). The idea is to fine-tune a second model with bounded mirrored gradient ascent, and to measure the normalized skew of this model with respect to the original one using consistent noise samples. We show that, while being conceptually simple and generic, MUCS systematically outperforms existing methods on three different datasets by a large margin. We additionally study the effect that core design choices have on final performance, and analyze novel aspects regarding the overlap of influential instances across generated items and the potential of ensembling TDA approaches. We believe that our findings may have broader implications for more general unlearning setups, as well as for tasks requiring the comparison of diffusion losses.
\end{abstract}


\section{Introduction}
\label{sec:intro}

In generative modeling, training data attribution (TDA)~\cite{hammoudeh_training_2024, deng_survey_2025} seeks to identify the training instances that causally influenced the generation of an item. Intuitively, such instances should be influential in a counterfactual sense~\cite{deng_survey_2025, georgiev_journey_2023, zheng_intriguing_2024, lin_diffusion_2024}, that is, removing them from training should prevent the model from generating the same item. This counterfactual notion of attribution is key to generative model interpretability~\cite{deng_survey_2025}, as well as to a myriad of downstream applications like data selection, leakage estimation, or memorization analysis~\cite{sun_enhancing_2025, ko_mirrored_2024, wang_data_2024}. Furthermore, there are domains in which attribution has direct practical value. For instance, in the music domain, TDA algorithms could inform the distribution of royalties derived from the generated samples~\cite{deng_computational_2024, morreale_attribution-by-design_2025, kim_generation_2025}. All these applications, however, still lack widespread adoption due to the limited performance and poor generalization of current TDA algorithms. Besides, proper counterfactual evaluation is also computationally demanding~\cite{deng_survey_2025, wang_data_2024}.

Diffusion models~\cite{lai_principles_2025} are the most prominent generative paradigm in domains like images or audio. This has motivated the adaptation of previous TDA algorithms to such paradigm~\cite{georgiev_journey_2023, zheng_intriguing_2024, lin_diffusion_2024}. However, with the introduction of more sophisticated diffusion variants~\cite{lai_principles_2025}, the adaptation of those TDA algorithms can become difficult or infeasible. On the other hand, there are a few non-parametric or model-agnostic methods, typically leveraging some form of embedding-based similarity~\cite{sun_enhancing_2025, zhao_nonparametric_2025}, but those can fall short in terms of performance and do not embody any notion of causality (see also the baselines in, e.g.,~\cite{georgiev_journey_2023, zheng_intriguing_2024, lin_diffusion_2024}). Unlearning is a relatively recent and general approach that has shown promising results for diffusion models, especially since the introduction of mirrored unlearning\footnote{Also called the ``mirrored influence hypothesis''~\cite{ko_mirrored_2024} or ``unlearning the synthesized image''~\cite{wang_data_2024}.}~\cite{ko_mirrored_2024, wang_data_2024, choi_large-scale_2025}. Its core idea is that unlearning a generated item should measurably affect the model's performance on the influential training instances. However, existing unlearning-based approaches rely on approximations, involve sensitive hyperparameters, offer limited control over the unlearning process, and employ simplistic attribution-scoring methods.

In this work, we introduce a method for TDA that combines a redefined mirrored unlearning process together with improvements in measuring the effect induced by such process. In contrast to existing works, our mirrored unlearning process combines a retention and a forgetting loss, and adds control by defining a null performance target (Sec.~\ref{sec:MUCS_lnull}) that is used both for preventing unnecessary unlearning and for stopping the unlearning process (Sec.~\ref{sec:MUCS_unlearn}). Our method also significantly differs from existing works in the way it performs attribution scoring (Sec.~\ref{sec:MUCS_score}): instead of relying on simple subtractions between loss expectations using the training noise distribution, it averages individual noise-consistent normalized loss skews using the generation noise distribution. We not only show that these components independently yield important performance improvements, but also that the method as a whole clearly outperforms existing TDA approaches (Sec.~\ref{sec:results}). We also conduct an analysis of attributed item overlap, and study the performance of attribution ensembles. Overall, our work introduces a number of important findings for TDA, both in terms of mirrored unlearning and attribution scoring.


\section{Method}
\label{sec:MUCS}

\subsection{Notation and Overview}
\label{sec:MUCS_overview}

Our method targets diffusion models trained under common principles~\cite{lai_principles_2025}, and is not specifically tied to individual network architectures, loss functions, diffusion variants, noise schedules, sampling schemes, or data types. As such, it only assumes a diffusion model $F_1$, pre-trained using some reconstruction loss $L$ leveraging noise scales $\sigma$, noise vectors $\ve{n}$, and training data instances $\ve{z}$. The latter are formed by data items $\ve{x}$ and, possibly, conditioning signals $\ve{c}$, such that $\ve{z}=\{\ve{x},\ve{c}\}$. We denote the training dataset by $\se{Z}=\{\ve{z}_1,\dots\ve{z}_{|\se{Z}|}\}$. At inference time, $F_1$ generates an item $\hat{\ve{x}}$ using conditioning $\hat{\ve{c}}$, forming $\hat{\ve{z}}=\{\hat{\ve{x}},\hat{\ve{c}}\}$. For the sake of generalization, we make $L$ implement the diffusion variant. For example, for DDPM~\cite{ho_denoising_2020},
\begin{equation*}
L(\ve{z},\sigma,\ve{n},F) = \mo{E}_{\sigma,\ve{n},\ve{z}}\left[ \norm{ \,\ve{n} - F\left(\sqrt{\alpha(\sigma)}\,\ve{x}+\sqrt{1-\alpha(\sigma)}\,\ve{n},\sigma,\ve{c}\right) }^2 \right]
\end{equation*}
and, for EDM~\cite{karras_elucidating_2022},
\begin{equation*}
L(\ve{z},\sigma,\ve{n},F) = \mo{E}_{\sigma,\ve{n},\ve{z}}\left[ \omega(\sigma) \norm{ \,\ve{x} - F\left(\ve{x}+\sigma\ve{n},\sigma,\ve{c}\right) }^2 \right] ,
\end{equation*}
where $\alpha$ and $\omega$ are weighting functions that depend on $\sigma$ (see~\cite{lai_principles_2025} for further diffusion variants). 

Given $F_1$, $L$, and $\hat{\ve{z}}$, our method computes an attribution score $a_i$ for each training instance $\ve{z}_i\in\se{Z}$, reflecting the degree to which $\ve{z}_i$ has contributed to (or been influential in) generating $\hat{\ve{x}}$:
\begin{equation*}
a_i=A(\ve{z}_i|\hat{\ve{z}},L,F_1) ,
\end{equation*}
where $A$ denotes the TDA algorithm. The computation of $a$ involves two equally-important steps: unlearning $\hat{\ve{z}}$ to obtain an unlearned model $F_2$ (Sec.~\ref{sec:MUCS_unlearn}) and averaging noise-consistent loss skews to obtain the final score $a$ (Sec.~\ref{sec:MUCS_score}). We term our method MUCS, for \underline{m}irrored \underline{u}nlearning and noise-\underline{c}onsistent loss \underline{s}kew. A one-page pseudo-code of the full method is provided in Appendix~\ref{sec:app_method_code}. The code we used to run our experiments is released at \link{}.

\subsection{Null Loss Reference}
\label{sec:MUCS_lnull}

Before running MUCS, we estimate a null loss value which will serve as an upper bound reference for the random (or null) performance of $F$. For that, we use the randomly-initialized model weights, which we denote by $F_0$, and estimate
\begin{equation}
L_\tx{null} = \mo{E}_{\sigma,\ve{n},\ve{z}} \left[ L(\ve{z},\sigma,\ve{n},F_0) \right] ,
\label{eq:lnull}
\end{equation}
where $\sigma$ and $\ve{n}$ are drawn from the same distributions used in the training of $F_1$, and $\ve{z}$ is sampled uniformly without replacement from $\se{Z}$. In Eq.~\ref{eq:lnull}, the expectation can be computed from a few thousands of realizations or until we run out of data. We note that, alternatively, given normalized training data, $L_\tx{null}$ can be analytically derived for some common diffusion loss functions\footnote{For example, given image data between $-$1 and 1, the $L_\tx{null}$ for a well-calibrated EDM approach~\cite{karras_elucidating_2022} is 1.}~\cite{lai_principles_2025}.

\subsection{Unlearning Approach}
\label{sec:MUCS_unlearn}

Given a generated item, MUCS proceeds by unlearning it as if it was part of the training set. This corresponds to the so-called mirrored unlearning paradigm~\cite{ko_mirrored_2024, wang_data_2024, choi_large-scale_2025}, which assumes that unlearning a generated item $\hat{\ve{z}}$ will impact the model's performance on the training instances $\ve{z}_i$ that contributed to such generation. While this does not establish a causal relation from $\ve{z}_i$ to $\hat{\ve{z}}$, Ko~et~al.~\cite{ko_mirrored_2024} and Wang~et~al.~\cite{wang_data_2024} empirically show that it significantly correlates with the impact that unlearning $\ve{z}_i$ has on $\hat{\ve{z}}$. Such correlation and impact can be measured directly using $L$, and exploited for TDA and related tasks (Sec.~\ref{sec:intro}). 
Instead of performing direct gradient ascent on $\hat{\ve{z}}$~\cite{ko_mirrored_2024}, or constraining it with the Fisher information matrix~\cite{wang_data_2024, choi_large-scale_2025}, we propose to employ gradient ascent as a regularization to fine-tuning with training data. 

Following mirrored unlearning, we aim to obtain a model $F_2$ that has unlearned the generated sample $\hat{\ve{z}}$. To do so, we copy the pre-trained model $F_1$ and perform fine-tuning (FT) with the train set $\se{Z}$ and, at the same time, mirrored gradient ascent (GA) using $\hat{\ve{z}}$. Specifically, we iteratively minimize
\vspace{0.1cm}
\begin{equation}
\begin{split}
L_\tx{MUCS}\left(\se{Z},\hat{\ve{z}},L,F_2\right) & = L_\tx{FT}\left(\se{Z},L,F_2\right) - \lambda\, L_\tx{GA}\left(\hat{\ve{z}},L,F_2\right) \\
   & = \mo{E}_{\sigma,\ve{n},\ve{z}}\left[ \,L\left(\ve{z},\sigma,\ve{n},F_2\right)\, \right] - \lambda\, \mo{E}_{\sigma,\ve{n}}\left[ \,\min\left( L\left(\hat{\ve{z}},\sigma,\ve{n},F_2\right) , L_\tx{null} \right)\, \right] ,
\end{split}
\label{eq:MUCS_loss}
\end{equation}
where $\sigma$ and $\ve{n}$ are drawn from the same distributions used in the training of $F_1$, $\ve{z}$ is sampled uniformly without replacement from $\se{Z}$, and $\lambda>0$ is a hyperparameter that controls the amount of unlearning (notice the negative sign). In Eq.~\ref{eq:MUCS_loss}, expectations are estimated from batches containing 100~independent draws of $\sigma$ and $\ve{n}$ (and $\ve{z}$). 
In preliminary experiments, we observed that performance was largely insensitive to the choice of batch size.

\topic{Considerations} Having defined $L_\tx{MUCS}$, we now make a number of considerations for better unlearning, and later validate them on the attribution task with an ablation study (Sec.~\ref{sec:results}).

{\renewcommand\labelenumi{C\theenumi.}
\begin{enumerate}

\item First of all, note that the joint optimization of $L_\tx{FT}$ and $L_\tx{GA}$ will encourage a balanced unlearning step where training data information is preserved but, at the same time, information related to $\hat{\ve{z}}$ is destroyed. The hyperparameter $\lambda$ will control such balance, and we typically want to constrain it to the range $\lambda\in(0,1]$ to prevent an overly aggressive unlearning. After some preliminary experiments, we set $\lambda=0.2$. 

\item One important consideration for unlearning algorithms is how to control the amount of unlearning, that is, how to produce a meaningful unlearning effect while not destroying the overall generation performance. Here, 
we employ a twofold strategy: we control the amount of unlearning by clamping to $L_\tx{null}$ (Eq.~\ref{eq:MUCS_loss}) and stop updating the model once $L_\tx{GA}$ reaches a value close to random or null performance (e.g.,~$L_\tx{GA}\geq 0.95\cdot L_\tx{null}$). Using the former prevents pushing the unlearned model towards extreme loss values, avoiding unrealistic unlearning. Using the latter allows for more dynamic unlearning processes which depend on how difficult it is to push the generated sample $\hat{\ve{z}}$ towards null model performance, avoiding insufficient or excessive unlearning steps on a case-by-case basis. 

\item We believe that employing the same configuration as in pre-training is also important for robust unlearning. Hence, to minimize $L_\tx{MUCS}$, we employ the same noise schedule, loss function, optimizer type, and hyperparameters used in the pre-training of $F_1$. The only exception to the above rule is the learning rate, which can actually vary during pre-training. By default, we set it constant and one order of magnitude below the nominal pre-training value. 

\item Finally, as it has been reported that unlearning only certain parts of the model can yield some improvement for TDA~\cite{wang_data_2024}, we perform unlearning only on the MLP weights of $F_2$ (both in the Transformer and in the conditioning network, see Appendix~\ref{sec:app_method_model}). Although in our case we do not expect a huge difference in performance, unlearning only certain blocks or certain parts of the model is customary in fine-tuning tasks~\cite{kumari_multi-concept_2023}, as well as in generic unlearning setups~\cite{kumari_ablating_2023}.

\end{enumerate}
}

\subsection{Attribution Score}
\label{sec:MUCS_score}

With the unlearned model $F_2$, we can now proceed to attribute (or score the influence of) the training instances $\ve{z}_i$. To do so, we first form a set of pairs of noise scales $\sigma$ and noise vectors $\ve{n}$,
$\se{N}=\{(\sigma_1,\ve{n}_1),\dots(\sigma_{|\se{N}|},\ve{n}_{|\se{N}|})\}$. 
We do it by taking $\sigma$ from the schedule used during generation and drawing $\ve{n}$ from the same distribution used during training. Then, for each training instance $\ve{z}_i\in\se{Z}$, we compute the normalized skew between the unlearned and the original model losses, and average across pairs:
\begin{equation}
A_\tx{MUCS}(\ve{z}_i|\hat{\ve{z}},L,F_1) = \frac{1}{|\se{N}|} \sum_{(\sigma_j,\ve{n}_j)\in\se{N}} \frac{ L\left(\ve{z}_i,\sigma_j,\ve{n}_j,F_2\right) - L\left(\ve{z}_i,\sigma_j,\ve{n}_j,F_1\right) }{ \vert L\left(\ve{z}_i,\sigma_j,\ve{n}_j,F_2\right) \vert + \vert L\left(\ve{z}_i,\sigma_j,\ve{n}_j,F_1\right) \vert + \epsilon } ~,
\label{eq:attrib_score}
\end{equation}
where $\epsilon>0$ is a small constant added for numerical stability (e.g.,~$\epsilon=10^{-3}$). We pre-sample $|\se{N}|=100$~pairs and, importantly, we reuse the same set $\se{N}$ for every $\ve{z}_i\in\se{Z}$. 
In preliminary experiments, we observed that increasing the size of $\se{N}$ slightly improves performance, but incurs extra computational cost.

\topic{Considerations} Having defined $A_\tx{MUCS}$, we now make a number of considerations with regard to the novelty of its components, and later assess their impact with our ablation study (Sec.~\ref{sec:results}).

{\renewcommand\labelenumi{C\theenumi.}
\begin{enumerate}

\item First of all, we note that $A_\tx{MUCS}$ utilizes a normalized skew formula, which differs from the common loss subtractions or dot products used in existing attribution approaches~\cite{hammoudeh_training_2024, deng_survey_2025}, including the unlearning-based ones~\cite{ko_mirrored_2024, wang_data_2024, choi_large-scale_2025}. We advocate for using normalized skews instead of subtractions or dot products in order to compensate for abnormally high or low losses $L_i=L(\ve{z}_i,\sigma,\ve{n},F_1)$. A normalized skew can be interpreted as a symmetric ratio\footnote{We also experimented with plain loss ratios, obtaining slightly worse and less numerically-stable results.} or a normalized subtraction, so that it produces a different effect depending on $L_i$ (e.g.,~it is different to have a loss difference of 0.1 if $L_i=0.1$ than if $L_i=0.5$). This way, any loss change induced by unlearning $\hat{\ve{z}}$ is measured in relation to the original loss.

\item Another consideration to make, and which also differs from existing TDA approaches, is the fact that we use noise-consistent measurements for computing $A_\tx{MUCS}$. More specifically, we compute all the loss terms in Eq.~\ref{eq:attrib_score} using the same gain value $\sigma_j$ and noise realization $\ve{n}_j$, and always calculate the average over the same set of elements $\se{N}$. That is, we employ the same $(\sigma_j,\ve{n}_j)$ in all operations, and the same $\se{N}$ for every $\ve{z}_i\in\se{Z}$. Without noise-consistency, the quantities in Eq.~\ref{eq:attrib_score} could be dominated by random fluctuations due to different noise realizations, as diffusion loss values can vary a lot with different $\sigma$ and, more importantly, with different $\ve{n}$. 

\item Finally, we find it relevant for $A_\tx{MUCS}$ to take the noise scales $\sigma$ from the schedule used during the generation of $\hat{\ve{x}}$, instead of drawing them from the distribution used during the training of $F_1$ (compare with $L_\tx{MUCS}$ in Eq.~\ref{eq:MUCS_loss}). In the image domain, it is well-known that large noise scales are more influential for general item properties such as object placement, spatial distribution, and overall color choices (see, e.g.,~\cite{bertrand_closed-form_2025}). By employing the generation schedule instead of the training distribution, which usually concentrates more on intermediate noise scales (see, e.g.,~\cite{karras_elucidating_2022}), we prime $A_\tx{MUCS}$ to focus more on those general properties. In addition, since it is generally observed that the last diffusion steps (lowest $\sigma$) are only responsible for small details and textures (see, e.g.,~\cite{georgiev_journey_2023}), we decide to discard the last 30\% of the schedule. Specifically, we obtain $\lfloor100/0.7\rfloor$ values for $\sigma$ following the generation schedule (in descending order), and include only the first 100 in $\se{N}$. 

\end{enumerate}
}


\section{Related Works}
\label{sec:relatedwork}

There are only a few works performing TDA in diffusion models via unlearning. In principle, unlearning every item in the training set is not practical due to the size of existing datasets, but the work of Ko~et~al.~\cite{ko_mirrored_2024} shifts this perspective with the mirrored influence approach. They perform unlearning with simple gradient ascent on the synthesized image. Concurrently, Wang~et~al.~\cite{wang_data_2024} also proposed to unlearn the synthesized image through gradient ascent. They introduce the Fisher information matrix as a constraint to prevent the unlearning of important weights, a technique borrowed from elastic weight consolidation (EWC)~\cite{kirkpatrick_overcoming_2017}. Choi~et~al.~\cite{choi_large-scale_2025} adapt the latter approach to the music domain and on a large scale. None of these approaches includes any fine-tuning notion nor employ loss skews. Moreover, they perform unbounded unlearning using a constant number of steps (or gradient accumulations), and compute loss subtractions using non-consistent noise realizations.

Besides unlearning, there are other strategies for TDA in diffusion models. The most common (albeit computationally demanding) one is to consider the notion of influence functions~\cite{koh_understanding_2017}, and to develop it for specific diffusion variants. Georgiev~et~al.~\cite{georgiev_journey_2023} and Zheng~et~al.~\cite{zheng_intriguing_2024} do that for DDPM, incorporating the use of random projections to make gradient and Hessian calculations tractable, following TRAK~\cite{park_trak_2023}. Lin~et~al.~\cite{lin_diffusion_2024} go one step further and introduce vector normalization and the KL divergence into the scoring procedure, obtaining state-of-the-art results. There are also several approaches that do not exploit the notion of influence functions (e.g.,~\cite{dai_ablation_2024, brokman_montrage_2024, mlodozeniec_distributional_2025}), and some recent methods that attempt to perform fast, embedding-based attribution (e.g.,~\cite{sun_enhancing_2025, zhao_nonparametric_2025, wang_fast_2025}). The latter may however be model- and/or domain-specific. For a general introduction to TDA, we refer the reader to the surveys by Hammoudeh~and~Lowd~\cite{hammoudeh_training_2024} and Deng~et~al.~\cite{deng_survey_2025}. To the best of our knowledge, previous TDA approaches do not consider the use of a normalized loss skew, noise-consistent realizations, and inference schedules for score calculation. Many TDA approaches also focus on specific diffusion variants like DDPM, which sometimes complicates their adaptation to more modern developments like EDM or flow matching~\cite{lai_principles_2025}.

To put our unlearning contribution into perspective, it is also worth to review existing unlearning approaches that are not tailored to TDA or diffusion models. Like ours, Alberti~et~al.'s approach~\cite{alberti_data_2025} can be understood to perform both fine-tuning and gradient ascent on selected samples. Similarly, Heng~and~Soh~\cite{heng_selective_2023} consider fine-tuning and gradient ascent for continual learning, but constrain the latter with EWC. Both of these works do not explicitly consider reusing the original model configuration, employ a constant number of unlearning steps, and do not leverage any notion of random performance or a null loss. In general, unlearning methods employ alternative optimization objectives. For example, Golatkar~et~al.~\cite{golatkar_eternal_2020} employ a KL divergence objective instead of plain gradient ascent, Grandikota~et~al.~\cite{gandikota_erasing_2023} use the predictions of the original (frozen) model instead of the available ground truth, Wu~et~al.~\cite{wu_erasing_2025} replace targets with randomly sampled instances, 
and Zhang~et~al.~\cite{zhang_defensive_2024} and Shi~et~al.~\cite{shi_retrack_2025} employ strategies inspired by classifier-free guidance~\cite{ho_classifier-free_2021}. Additional approaches with similar considerations have been proposed for continual learning~\cite{liu_continual_2022}, preference optimization~\cite{zhang_negative_2024}, classifiers~\cite{tarun_fast_2024, kurmanji_towards_2023}, or large language models~\cite{veldanda_llm_2024, ren_general_2025}. Overall, we could not find an unlearning approach that matches most of the aspects of our unlearning formulation.


\section{Evaluation Methodology}
\label{sec:eval}

\topic{Data} 
To demonstrate generalization with respect to different types of images, resolutions, and conditioning signals, we consider three complementary datasets: CIFAR10~\cite{krizhevsky_learning_2009}, ArtBench10~\cite{liao_artbench_2022}, and the captioned version of MS-COCO~\cite{fang_captions_2015}, which we denote as COCO. For CIFAR10 we use a resolution of 32$\times$32, while for ArtBench10 and COCO we use a resolution of 64$\times$64. To simulate different scenarios, we do not employ any conditioning for CIFAR10, condition on the class index for ArtBench10, and use the CLIP-Text~\cite{radford_learning_2021} embeddings of the original captions for COCO. For all datasets, we use the full original train/test splits. Although these datasets are routinely used in prior work, they are seldom considered jointly, and evaluations are often limited to a few classes (e.g.,~two classes from CIFAR10 or two classes from ArtBench10). Besides, very few TDA approaches consider conditional diffusion models. Although we are still far from the real-world datasets used to train current large-scale diffusion models, this setup substantially improves over existing approaches and pushes the TDA evaluation standards towards a more realistic scenario.

\topic{Diffusion Model} 
Our models follow a typical diffusion pipeline based on the DiT/B architecture~\cite{peebles_scalable_2023} and the EDM variant~\cite{karras_elucidating_2022}. For computational and practical reasons, the model works in the pixel space, and uses a two-layer convolutional encoder and decoder which are jointly trained with the main Transformer blocks. For conditional generation, we employ condition dropout during training and classifier-free guidance~\cite{ho_classifier-free_2021} during inference. Weights are initialized using PyTorch's~\cite{paszke_pytorch_2019} defaults, and we optimize them with AdamW~\cite{loshchilov_decoupled_2019}. In preliminary work, we conducted a few experiments with the DiT/L and UNet architectures, and also with the DDPM variant~\cite{ho_denoising_2020}, obtaining equivalent results (not reported due to computational constraints). Appendix~\ref{sec:app_method_model} provides additional specifications of the training and generation procedures, together with the configuration differences for each dataset.

\topic{Baselines} 
To fairly compare with existing methods, we re-implement several baseline approaches, which we group under three broad categories: model-agnostic, influence function-based, and unlearning-based. Model-agnostic approaches rely on existing similarity measures to derive attribution scores. We consider cosine similarity of the condition vector (Condition; one-hot for CIFAR10/ArtBench10, CLIP-Text for COCO), cosine similarity in the CLIP-Image space (CLIP)~\cite{radford_learning_2021}, and cosine similarity in the DINOv2 space (DINO)~\cite{oquab_dinov2_2023}. Influence function-based approaches rely on gradient and Hessian calculations from multiple noise samples. We consider two of the most competitive approaches in this area: diffusion TRAK (D-TRAK)~\cite{zheng_intriguing_2024} and diffusion attribution score~(DAS)~\cite{lin_diffusion_2024}. For computational reasons, we project gradients to 16384 dimensions (preliminary experiments with a larger dimensionality were unstable and did not yield a dramatic improvement), and fine-tune the other hyperparameters on each dataset. Unlearning-based approaches leverage some techniques like the ones discussed in this paper. We consider the two existing approaches: the mirrored influence hypothesis approach (Forward-INF)~\cite{ko_mirrored_2024} and the Hessian-weighted approach of attribution by unlearning (AbU)~\cite{wang_data_2024}. We fine-tune their hyperparameters on the CIFAR10 dataset. All approaches that consider batches of noise samples use 100~samples like ours. 

\begin{figure}[t]
\centering
\includegraphics[width=0.95\linewidth]{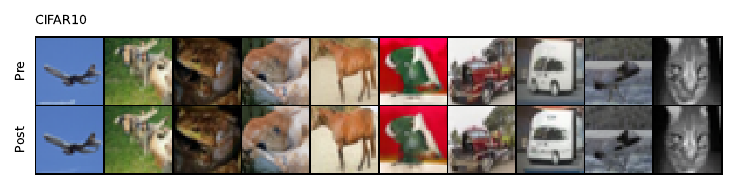}
\includegraphics[width=0.95\linewidth]{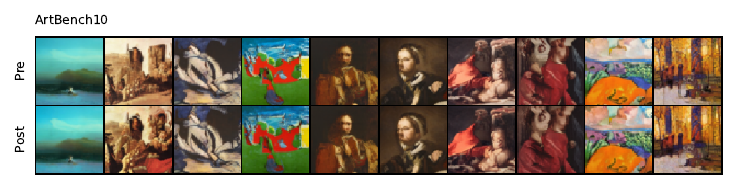}
\includegraphics[width=0.95\linewidth]{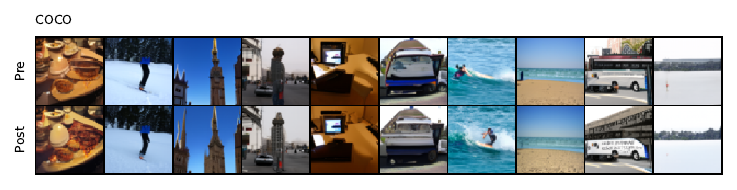}
\caption{Examples of images generated using the original pre-trained model (Pre) and the post-trained model with 40\% random data removal (Post). We can observe the strikingly high similarity for images generated with the same seed (columns), despite using a model retrained from scratch with a considerably smaller subset of the original data. }
\label{fig:image_pairs_random}
\end{figure}

\topic{Attribution Performance} 
We evaluate TDA using a leave-$k$-out counterfactual setup~\cite{deng_survey_2025, wang_data_2024} (see also~\cite{georgiev_journey_2023, zheng_intriguing_2024, lin_diffusion_2024}; considerations about alternative evaluation approaches are given in Appendix~\ref{sec:app_method_eval}). Essentially, this setup tests the ability of a model to generate a previously generated image after retraining from scratch without including its topmost $k$ influential items in the training set. To quantify this ability, we rely on the seed-consistency property of diffusion models~\cite{georgiev_journey_2023, wang_data_2024}, which implies that randomly removing non-influential items and retraining yields generations that are strikingly similar to the ones produced by the original model using the same seed (see Fig.~\ref{fig:image_pairs_random} for examples, and also~\cite{georgiev_journey_2023, wang_data_2024}). To assess image similarity, we employ four measures covering multiple and complementary aspects: the structural similarity index measure (SSIM)~\cite{wang_image_2004}, cosine similarity in the SSCD copy-detection space (SSCD)~\cite{pizzi_self-supervised_2022}, the learned perceptual image patch similarity (LPIPS) measure~\cite{zhang_unreasonable_2018}, and cosine similarity in the semantic CLIP space (CLIP)~\cite{radford_learning_2021}.

The main idea behind our evaluation is that good TDA algorithms should select training items that, once removed from the training set, significantly alter the generations of the retrained model (compared to removing items at random, using the same seed). Given a set of $m$ items $\hat{\se{Z}}=\{\hat{\ve{z}}_1,\dots\hat{\ve{z}}_m\}$ generated using seeds $\se{U}=\{u_1,\dots u_m\}$, we compute the top-$k$ most influential training items for each $\hat{\ve{z}}_i\in\hat{\se{Z}}$ using some attribution method $A$. Next, we remove those items from the train set ($mk$ removals in total) and retrain the diffusion model from scratch. We then generate a new set of items $\hat{\se{Z}}'=\{\hat{\ve{z}}'_1,\dots\hat{\ve{z}}'_m\}$ using the same seeds $\se{U}$, and compare their paired image similarities $\se{S}=\{s_1,\dots s_m\}$, $s_i=S(\hat{\ve{x}}_i,\hat{\ve{x}}'_i)$, where $S$ corresponds to the similarity metrics above. We perform this process for every attribution algorithm $A$ and a random null model $A_\tx{rand}$ that simply attributes items at random. 

\begin{figure}[t]
\centering
\includegraphics[width=0.9\linewidth]{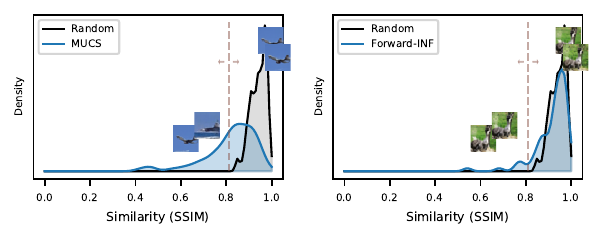}
\vspace{-0.2cm}
\caption{Examples of the distributions of similarities between $\ve{z}$ and $\ve{z}'$ for models trained on CIFAR10. Black and blue colors correspond to $A_\tx{rand}$ and a specific method $A$, respectively (see text). Better attribution algorithms (left) present less overlap than worse ones (right).}
\label{fig:example_distribs}
\end{figure}

To assess the degree of change produced by $A$ with respect to $A_\tx{rand}$, we measure the overlap between similarity distributions (Fig.~\ref{fig:example_distribs}). To objectively quantify such overlap, we use the area under the ROC curve (AUC), which is a common evaluation measure employed for that purpose~\cite{bradley_use_1997}.
The AUC also has the nice property of being proportional to the U statistic of the Mann-Whitney~U test for measuring whether two samples correspond to the same distribution~\cite{mason_areas_2002}, and we leverage that to compute statistical significance (one tail, $p<0.05$). Besides AUCs, we also report average similarity differences and their confidence intervals, so that we obtain a complementary perspective on distribution overlap (Appendix~\ref{sec:app_results_compare}). Our final setup involves removing $k=2$\% of the training data for $m=20$ generations (totaling a maximum of 40\% of the training data) and running it 6~times using different network and seed initializations. Hence, evaluations are made over 120~generated items, which is similar to or more than previous leave-$k$-out counterfactual evaluations (cf.~\cite{georgiev_journey_2023, zheng_intriguing_2024, lin_diffusion_2024, wang_data_2024}). A more elaborate description of our evaluation approach can be found in Appendix~\ref{sec:app_method_eval}.

\section{Results}
\label{sec:results}

We start by comparing MUCS with existing approaches (Table~\ref{tab:comparison}). We see that MUCS outperforms all considered approaches in all similarity metrics by a large margin. The only exception among the 12~considered cases is the CLIP metric in the COCO dataset. The difference with the best-performing baseline is statistically significant for almost all metrics in both CIFAR10 and ArtBench10 datasets. For COCO, the difference is not as significant, yet still substantial (COCO is a more complex dataset, and statistical significance is hard to obtain unless the baseline performs poorly). Besides MUCS, it is interesting to note that D-TRAK and DAS consistently perform better than model-agnostic or the other unlearning approaches. Model-agnostic approaches like CLIP or DINO can be very competitive in some datasets and metrics (e.g.,~CLIP in COCO). However, we see that their performance on other datasets can be relatively poor (e.g.,~CIFAR10 and ArtBench10). Previous unlearning-based methods do not perform well in general, perhaps due to their suboptimal formulation and the difficulty to setup some of their hyperparameters. MUCS overcomes this situation with significant improvement, while also being 1.5--2.9~times faster than the considered influence function- and unlearning-based approaches. This and additional results are available in Appendix~\ref{sec:app_results_compare}, together with image regeneration examples.

\begin{table}[t]
  \caption{Comparison with existing approaches, which can be categorized into model-agnostic (MA), influence function-based (IF), and unlearning-based (U). We report AUC values for the considered similarity measures, plus their average ($\mu$; last column). 
  The asterisk (*) denotes statistical significance with respect to the closest baseline (Mann-Whitney U; $p<0.05$).}
  \label{tab:comparison}
  \centering
  \resizebox{0.94\columnwidth}{!}{%
  \begin{tabular}{llcccccc}
    \toprule
     & \textbf{Approach} & \textbf{Categ.} 	& \textbf{SSIM}& \textbf{SSCD} & \textbf{LPIPS} & \textbf{CLIP} & $\bm\mu$ \\
    \midrule
    \multirow{8}{*}{\rotatebox[origin=c]{90}{CIFAR10}} 
    & Condition 							& MA & 0.572 & 0.576 & 0.568 & 0.550 & 0.566 \\
    & CLIP	 								& MA & 0.650 & 0.672 & 0.672 & \cellcolor{bronze}0.646 & 0.660 \\
    & DINO	 					            & MA & 0.633 & 0.635 & 0.648 & 0.605 & 0.630 \\
    & D-TRAK~\cite{zheng_intriguing_2024}	& IF & \cellcolor{bronze}0.749 & \cellcolor{bronze}0.702 & \cellcolor{bronze}0.736 & 0.621 & \cellcolor{bronze}0.702 \\
    & DAS~\cite{lin_diffusion_2024} 		& IF & \cellcolor{silver}\fu{0.810} & \cellcolor{silver}\fu{0.787} & \cellcolor{silver}\fu{0.813} & \cellcolor{silver}\fu{0.735} & \cellcolor{silver}\fu{0.786} \\
    & Forward-INF~\cite{ko_mirrored_2024} 	& U  & 0.570 & 0.523 & 0.554 & 0.507 & 0.539 \\
    & AbU~\cite{wang_data_2024} 			& U  & 0.523 & 0.556 & 0.537 & 0.576 & 0.548 \\
    & MUCS (proposed)    					& U  & \cellcolor{gold}~~\fb{0.889}* & \cellcolor{gold}~~\fb{0.901}* & \cellcolor{gold}~~\fb{0.930}* & \cellcolor{gold}~~\fb{0.868}* & \cellcolor{gold}\fb{0.897} \\
    \midrule
    \multirow{8}{*}{\rotatebox[origin=c]{90}{ArtBench10}} 
    & Condition 							& MA & 0.547 & 0.542 & 0.547 & 0.560 & 0.549 \\
    & CLIP	 								& MA & 0.641 & 0.643 & 0.648 & 0.685 & 0.654 \\
    & DINO	 					            & MA & 0.565 & 0.573 & 0.579 & 0.586 & 0.576 \\
    & D-TRAK~\cite{zheng_intriguing_2024}	& IF & \cellcolor{bronze}0.769 & \cellcolor{bronze}0.744 & \cellcolor{bronze}0.764 & \cellcolor{bronze}0.715 & \cellcolor{bronze}0.748 \\
    & DAS~\cite{lin_diffusion_2024} 		& IF & \cellcolor{silver}\fu{0.783} & \cellcolor{silver}\fu{0.762} & \cellcolor{silver}\fu{0.782} & \cellcolor{silver}\fu{0.722} & \cellcolor{silver}\fu{0.762} \\
    & Forward-INF~\cite{ko_mirrored_2024} 	& U  & 0.669 & 0.627 & 0.650 & 0.623 & 0.642 \\
    & AbU~\cite{wang_data_2024} 			& U  & 0.503 & 0.530 & 0.503 & 0.505 & 0.510 \\
    & MUCS (proposed)    					& U  & \cellcolor{gold}\fb{0.850} & \cellcolor{gold}~~\fb{0.840}* & \cellcolor{gold}~~\fb{0.859}* & \cellcolor{gold}~~\fb{0.823}* & \cellcolor{gold}\fb{0.843} \\
    \midrule
    \multirow{8}{*}{\rotatebox[origin=c]{90}{COCO}} 
    & Condition 							& MA & 0.610 & 0.645 & 0.630 & 0.633 & 0.629 \\
    & CLIP	 								& MA & 0.678 & 0.705 & 0.719 & \cellcolor{silver}\fu{0.764} & \cellcolor{bronze}0.716 \\
    & DINO	 					            & MA & 0.682 & 0.720 & 0.710 & \cellcolor{gold}\fb{0.766} & \cellcolor{silver}\fu{0.720} \\
    & D-TRAK~\cite{zheng_intriguing_2024}	& IF & \cellcolor{bronze}0.720 & \cellcolor{bronze}0.708 & \cellcolor{bronze}0.742 & 0.683 & 0.713 \\
    & DAS~\cite{lin_diffusion_2024} 		& IF & \cellcolor{silver}\fu{0.726} & \cellcolor{silver}\fu{0.713} & \cellcolor{silver}\fu{0.743} & 0.680 & 0.715 \\
    & Forward-INF~\cite{ko_mirrored_2024} 	& U  & 0.616 & 0.620 & 0.634 & 0.624 & 0.624 \\
    & AbU~\cite{wang_data_2024} 			& U  & 0.531 & 0.523 & 0.543 & 0.526 & 0.531 \\
    & MUCS (proposed)    					& U  & \cellcolor{gold}\fb{0.750} & \cellcolor{gold}\fb{0.763} & \cellcolor{gold}~~\fb{0.799}* & \cellcolor{bronze}0.715 & \cellcolor{gold}\fb{0.757} \\
   \bottomrule
  \end{tabular}
  }
\end{table}

Next, we turn our attention to the specific MUCS components and their effect on overall performance (Table~\ref{tab:ablation}). We first observe that the hyperparameter $\lambda$ has a mild effect when it is kept close to the chosen value of 0.2, with a global performance variation around 3\% on COCO (U--C1). Sampling $\sigma$ values from a slightly modified distribution during unlearning has a similar effect (U--C2), but we note that it can potentially be beneficial for some metrics (see CLIP; we leave this aspect open for future research). Employing the concept of $L_\tx{null}$ for both preventing excessive unlearning and deciding when to stop proves important, with a performance drop above 4\% when using the MUCS average number of performed unlearning steps (U--C3). Unlearning the full Transformer block yields a performance drop around 5\%. It is however in the part concerning attribution score calculation where our design choices have the most impact (Table~\ref{tab:ablation}, bottom). Substituting our normalized loss skew formulation by simple loss subtraction yields a performance drop of about 6\% (S--C1), and using the training distribution for sampling $\sigma$ causes another 5\% drop (S--C3). Using non-consistent noises has the highest impact, with AUC scores dropping around 19\% on average (S--C2). We hope that these design choices can be adopted by other TDA approaches beyond unlearning-based ones.

\begin{table}[t]
  \caption{Ablation study on COCO. ``Abl.'' refers to the MUCS step and consideration where the ablation is applied (e.g.,~U--C1 denotes the unlearning step, consideration~1; S--C2 refers to the scoring step, consideration~2). ``Description'' summarizes what changes with respect to the proposed approach (see also main text). Numbers correspond to AUC values for the considered similarity measures, plus their average relative difference ($\Delta$; last column). 
  }
  \label{tab:ablation}
  \centering
  \resizebox{0.98\columnwidth}{!}{%
  \begin{tabular}{clccccc}
    \toprule
    \textbf{Abl.} & \textbf{Description} 	& \textbf{SSIM}& \textbf{SSCD} & \textbf{LPIPS} & \textbf{CLIP} & $\bm\Delta$ \\
    \midrule
    	 & MUCS (proposed)					& \fb{0.750} & \fb{0.763} & \fb{0.799} & 0.715 & {\footnotesize (0.757)} \\
    \midrule
    U--C1 & Using $\lambda=0.1$ 			    & 0.733 & 0.737 & 0.767 & 0.695 & ~$-$3.2\% \\
    U--C1 & Using $\lambda=0.3$ 				& 0.733 & 0.735 & 0.782 & 0.690 & ~$-$2.9\% \\
    U--C2 & Different $\sigma$ distribution 	& 0.734 & 0.732 & 0.771 & \fb{0.716} & ~$-$2.5\% \\
    U--C3 & No $L_\tx{null}$ and 109 unlearning steps 	& 0.716 & \cellcolor{myorange}0.713 & 0.764 & 0.707 & ~$-$4.2\% \\
    U--C4 & Unlearning full Transformer block& 0.730 & 0.736 & 0.779 & \cellcolor{myorange}0.681 & ~$-$3.3\% \\
    \midrule
    S--C1 & Loss subtraction (no skew)	& \cellcolor{myorange}0.703 & 0.724 & \cellcolor{myorange}0.753 & \cellcolor{myorange}0.676 & \cellcolor{myorange}~$-$5.7\% \\
    S--C2 & Non-consistent noise pairs	 		& \cellcolor{myorange}0.618 & \cellcolor{myorange}0.621 & \cellcolor{myorange}0.619 & \cellcolor{myorange}0.596 & \cellcolor{myorange}$-$19.0\%\, \\
    S--C3 & Full $\sigma$ generation schedule   & 0.745 & 0.747 & 0.795 & 0.695 & ~$-$1.6\% \\
    S--C3 & Use of $\sigma$ training distribution & \cellcolor{myorange}0.709 & \cellcolor{myorange}0.708 & \cellcolor{myorange}0.761 & 0.692 & \cellcolor{myorange}~$-$5.3\% \\
    \bottomrule
  \end{tabular}
  }
\end{table}

To conclude, we complement our work with an analysis of the overlap between top-$k$ attributed instances and an exploration of TDA ensembles. In all our experiments, we observe that all considered approaches present some top-$k$ overlap between generated items (Appendix~\ref{sec:app_results_overlap}). Having an overlap implies that there are some training instances that are influential for multiple generations at the same time, and suggests the potential existence of some `hub' or `centrality' bias~\cite{radovanovic_hubs_2010}. However, the overlaps observed with our experiments are relatively low, and close to random choice. 
While this deserves further investigation, we believe the observed degree of overlap is not critical for the evaluation of the considered approaches. 
If, instead of studying the top-$k$ overlap obtained between generated items, we study the top-$k$ overlap between attribution approaches, we observe that almost none of them presents an overlap over 35\% with another approach (Appendix~\ref{sec:app_results_ensemble}). This fact suggests that perhaps an ensemble approach could provide a significant boost to the task of TDA (to our knowledge, such an ensemble approach has not been previously considered in the literature). To test this hypothesis, we implement and run a simple rank-based ensembling approach (Appendix~\ref{sec:rankensemble}) and compute AUC values under different weightings (Table~\ref{tab:ensemble}). Interestingly, we observe that ensembling has almost always a detrimental effect for datasets in which we already obtained good AUCs (e.g.,~CIFAR10 and ArtBench10), but can have a positive effect for datasets where AUCs were not so good (e.g.,~COCO). Of course, the cost of performing an ensemble increases with the cost of the individual methods being considered, but we observe that combining MUCS with cheap, model-agnostic attribution measures like DINO, CLIP, or Condition can provide a boost in some datasets (ensemble B, COCO). We also observe that removing MUCS from the ensemble has a notable detrimental effect (ensemble A~vs.~C and B~vs.~D). In future work, it would be interesting to study how and when to combine TDA approaches to form an ensemble, especially including computationally-cheap approaches like model-agnostic or non-parametric ones (Sec.~\ref{sec:relatedwork}).

\begin{table}[t]
  \caption{Ensemble results for the considered datasets. Only average AUC scores per dataset are reported ($\mu$). Letters denote the MUCS reference (R) and the ensemble name (A--D). 
  }
  \label{tab:ensemble}
  \centering
  \setlength{\tabcolsep}{5pt}
  \resizebox{0.98\columnwidth}{!}{%
  \begin{tabular}{ccccccccccc}
    \toprule
    \textbf{\#} & \multicolumn{7}{c}{\textbf{Weight in the Ensemble}} & \multicolumn{3}{c}{\textbf{AUC} $\bm\mu$} \\
    \cmidrule{9-11}
    & MUCS & DAS & D-TRAK & DINO & CLIP & Cond & F-INF & \textbf{CIFAR10} & \textbf{ArtBench10} & \textbf{COCO} \\
    \midrule
    R & 10 & 0 & 0 & 0 & 0 & 0 & 0 & \textbf{0.897} & 0.843 & 0.757 \\
    \midrule
    A & 10 & 7 & 6 & 5 & 5 & 3 & 2 & \cellcolor{myorange}0.834 & \textbf{0.859} & \textbf{0.801} \\
    B & 10 & 0 & 0 & 5 & 5 & 3 & 0 & \cellcolor{myorange}0.713 & \cellcolor{myorange}0.813 & \textbf{0.801} \\
    C & 0  & 7 & 6 & 5 & 5 & 3 & 2 & \cellcolor{myorange}0.782 & \cellcolor{myorange}0.810 & 0.762 \\
    D & 0  & 0 & 0 & 5 & 5 & 3 & 0 & \cellcolor{myorange}0.605 & \cellcolor{myorange}0.709 & \cellcolor{myorange}0.755 \\
    \bottomrule
  \end{tabular}
  }
\end{table}


\section{Conclusion}

This paper proposes core methodologies for performing TDA in diffusion models. On the one hand, we propose an improved mirrored unlearning approach based on fine-tuning and gradient ascent, featuring gradient ascent clamping and a dynamic stopping criterion. On the other hand, we propose several performance-critical changes to how attribution scores should be computed, including noise-consistency, normalized skews, and using part of the generation noise schedule. The combination of these core methodologies yields MUCS, a TDA approach that significantly outperforms existing approaches on counterfactual evaluations using three different datasets. We hope that the proposed methodologies can further impact more general unlearning setups and tasks where diffusion losses need to be compared.


\section*{Acknowledgements}

We thank Woosung Choi for his AbU code review, and Chieh-hsin Lai and Yuhta Takida for initial discussions before starting this project.

{
\ifdefined\anon
\small
\else
\fi
\bibliographystyle{unsrtnat}
\bibliography{paper_references}

@article{lai_principles_2025,
	title = {The principles of diffusion models},
	journal = {ArXiv: 2510.21890},
	author = {Lai, C.-H. and Song, Y. and Kim, D. and Mitsufuji, Y. and Ermon, S.},
	year = {2025},
}

@inproceedings{zheng_intriguing_2024,
	title = {Intriguing properties of data attribution on diffusion models},
	booktitle = {Proc. of the {Int}. {Conf}. on {Learning} {Representations} ({ICLR})},
	author = {Zheng, X. and Pang, T. and Du, C. and Jiang, J. and Lin, M.},
	year = {2024},
}

@inproceedings{lin_diffusion_2024,
	title = {Diffusion attribution score: evaluating training data influence in diffusion models},
	booktitle = {Proc. of the {Int}. {Conf}. on {Learning} {Representations} ({ICLR})},
	author = {Lin, J. and Tao, L. and Dong, M. and Xu, C.},
	year = {2024},
}

@inproceedings{ko_mirrored_2024,
	title = {The mirrored influence hypothesis: efficient data influence estimation by harnessing forward passes},
	booktitle = {Proc. of the {IEEE} {Conf}. on {Computer} {Vision} and {Pattern} {Recognition} ({CVPR})},
	author = {Ko, M. and Kang, F. and Shi, W. and Jin, M. and Yu, Z. and Jia, R.},
	year = {2024},
	pages = {26286--26295},
}

@incollection{wang_data_2024,
	title = {Data attribution for text-to-image models by unlearning synthesized images},
	volume = {37},
	booktitle = {Advances in {Neural} {Information} {Processing} {Systems} ({NeurIPS})},
	author = {Wang, S.-Y. and Hertzmann, A. and Efros, A. A. and Zhu, J.-Y. and Zhang, R.},
	year = {2024},
	pages = {4235--4266},
}

@inproceedings{alberti_data_2025,
	title = {Data unlearning in diffusion models},
	booktitle = {Proc. of the {Int}. {Conf}. on {Learning} {Representations} ({ICLR})},
	author = {Alberti, A. and Hasanaliyev, K. and Shah, M. and Ermon, S.},
	year = {2025},
}

@incollection{choi_large-scale_2025,
	series = {Creative {AI} {Track}},
	title = {Large-scale training data attribution for music generative models via unlearning},
	booktitle = {Advances in {Neural} {Information} {Processing} {Systems} ({NeurIPS})},
	author = {Choi, W. and Koo, J. and Cheuk, K. and Serrà, J. and Martínez-Ramírez, M. A. and Ikemiya, Y. and Murata, N. and Takida, Y. and Liao, W.-H. and Mitsufuji, Y.},
	year = {2025},
	pages = {in press},
}

@inproceedings{kumari_multi-concept_2023,
	title = {Multi-concept customization of text-to-image diffusion},
	booktitle = {Proc. of the {IEEE} {Conf}. on {Computer} {Vision} and {Pattern} {Recognition} ({CVPR})},
	author = {Kumari, N. and Zhang, B. and Zhang, R. and Shechtman, E. and Zhu, J.-Y.},
	year = {2023},
	pages = {1931--1941},
}

@inproceedings{kumari_ablating_2023,
	title = {Ablating concepts in text-to-image diffusion models},
	booktitle = {Proc. of the {IEEE} {Conf}. on {Computer} {Vision} and {Pattern} {Recognition} ({CVPR})},
	author = {Kumari, N. and Zhang, B. and Wang, S.-Y. and Shechtman, E. and Zhang, R. and Zhu, J.-Y.},
	year = {2023},
	pages = {22691--22702},
}

@incollection{karras_elucidating_2022,
	title = {Elucidating the design space of diffusion-based generative models},
	booktitle = {Advances in {Neural} {Information} {Processing} {Systems} ({NeurIPS})},
	author = {Karras, T. and Aittala, M. and Aila, T. and Laine, S.},
	year = {2022},
	pages = {26565--26577},
}

@article{deng_survey_2025,
	title = {A survey of data attribution: methods, applications, and evaluation in the era of generative {AI}},
	journal = {SSRN: 5451054},
	author = {Deng, J. and Hu, Y. and Hu, P. and Li, T.-W. and Liu, S. and Wang, J. T. and Ley, D. and Dai, Q. and Huang, B. and Huang, J. and Jiao, C. and Just, H. A. and Pan, Y. and Shen, J. and Tu, Y. and Wang, W. and Wang, X. and Zhang, S. and Zhang, S. and Jia, R. and Lakkaraju, H. and Peng, H. and Tang, W. and Xiong, C. and Zhao, J. and Tong, H. and Zhao, H. and Ma, Jiaqi W},
	year = {2025},
}

@incollection{bertrand_closed-form_2025,
	title = {On the closed-form of flow matching: generalization does not arise from target stochasticity},
	number = {in press},
	booktitle = {Advances in {Neural} {Information} {Processing} {Systems} ({NeurIPS})},
	author = {Bertrand, Q. and Gagneux, A. and Massias, M. and Emonet, R.},
	year = {2025},
}

@inproceedings{georgiev_journey_2023,
	title = {The journey, not the destination: how data guides diffusion models},
	booktitle = {Proc. of the {ICML} {Workshop} on {Challenges} in {Deployable} {Generative} {AI}},
	author = {Georgiev, K. and Vendrow, J. and Salman, H. and Park, S. M. and Madry, A.},
	year = {2023},
}

@article{kirkpatrick_overcoming_2017,
	title = {Overcoming catastrophic forgetting in neural networks},
	volume = {114},
	number = {13},
	journal = {Proceedings of the National Academy of Sciences},
	author = {Kirkpatrick, J. and Pascanu, R. and Rabinowitz, N. and Veness, J. and Desjardins, G. and Rusu, A. A. and Milan, K. and Quan, J. and Ramalho, T. and Grabska-Barwinska, A. and Hassabis, D. and Clopath, C. and Kumaran, D. and Hadsell, R.},
	year = {2017},
	pages = {3521--3526},
}

@inproceedings{koh_understanding_2017,
	title = {Understanding black-box predictions via influence functions},
	booktitle = {Proc. of the {Int}. {Conf}. on {Machine} {Learning} ({ICML})},
	author = {Koh, P. W. and Liang, P.},
	year = {2017},
	pages = {1885--1894},
}

@inproceedings{park_trak_2023,
	title = {{TRAK}: attributing model behavior at scale},
	booktitle = {Proc. of the {Int}. {Conf}. on {Machine} {Learning} ({ICML})},
	author = {Park, S. M. and Georgiev, K. and Ilyas, A. and Leclerc, G. and Madry, A.},
	year = {2023},
	pages = {27074--27113},
}

@inproceedings{brokman_montrage_2024,
	title = {{MONTRAGE}: monitoring training for attribution of generative diffusion models},
	booktitle = {Proc. of the {European} {Conf}. on {Computer} {Vision} ({ECCV})},
	author = {Brokman, J. and Hofman, O. and Vainshtein, R. and Giloni, A. and Shimizu, T. and Singh, I. and Rachmil, O. and Zolfi, A. and Shabtai, A. and Unno, Y. and Kojima, H.},
	year = {2024},
	pages = {1--17},
}

@article{dai_ablation_2024,
	title = {Ablation based counterfactuals},
	journal = {ArXiv: 2406.07908},
	author = {Dai, Zheng and Gifford, David K.},
	year = {2024},
}

@article{hammoudeh_training_2024,
	title = {Training data influence analysis and estimation: a survey},
	volume = {113},
	journal = {Machine Learning},
	author = {Hammoudeh, Z. and Lowd, D.},
	year = {2024},
	pages = {2351--2403},
}

@article{zhao_nonparametric_2025,
	title = {Nonparametric data attribution for diffusion models},
	journal = {ArXiv: 2510.14269},
	author = {Zhao, Y. and Du, C. and Zheng, X. and Pang, T. and Lin, M.},
	year = {2025},
}

@article{mlodozeniec_distributional_2025,
	title = {Distributional training data attribution},
	journal = {ArXiv: 2506.12965},
	author = {Mlodozeniec, B. and Reid, I. and Power, S. and Krueger, D. and Erdogdu, M. and Turner, R. E. and Grosse, R.},
	year = {2025},
}

@incollection{sun_enhancing_2025,
	title = {Enhancing training data attribution with representational optimization},
	booktitle = {Advances in {Neural} {Information} {Processing} {Systems} ({NeurIPS})},
	author = {Sun, W. and Liu, H. and Kandpal, N. and Raffel, C. and Yang, Y.},
	year = {2025},
	pages = {in press},
}

@article{wang_fast_2025,
	title = {Fast data attribution for text-to-image models},
	journal = {ArXiv: 2511.10721},
	author = {Wang, S.-Y. and Hertzmann, A. and Efros, A. A. and Zhang, R. and Zhu, J.-Y.},
	year = {2025},
}

@inproceedings{liu_continual_2022,
	title = {Continual learning and private unlearning},
	booktitle = {Proc. of the {Conf}. on {Lifelong} {Learning} {Agents} ({CoLLAs})},
	author = {Liu, B. and Stone, P.},
	year = {2022},
	pages = {243--254},
}

@inproceedings{zhang_negative_2024,
	title = {Negative preference optimization: from catastrophic collapse to effective unlearning},
	booktitle = {Proc. of the {Conf}. on {Language} {Modeling} ({COLM})},
	author = {Zhang, R. and Lin, L. and Bai, Y. and Mei, S.},
	year = {2024},
}

@article{tarun_fast_2024,
	title = {Fast yet effective machine unlearning},
	volume = {35},
	number = {9},
	journal = {IEEE Transactions on Neural Networks and Learning Systems},
	author = {Tarun, A. K. and Chundawat, V. S. and Mandal, M. and Kankanhalli, M.},
	year = {2024},
	pages = {13046--13055},
}

@incollection{kurmanji_towards_2023,
	title = {Towards unbounded machine unlearning},
	booktitle = {Advances in {Neural} {Information} {Processing} {Systems} ({NeurIPS})},
	author = {Kurmanji, M. and Triantafillou, P. and Hayes, J. and Triantafillou, E.},
	year = {2023},
	pages = {1957--1987},
}

@article{veldanda_llm_2024,
	title = {{LLM} surgery: efficient knowledge unlearning and editing in large language models},
	journal = {ArXiv: 2409.13054},
	author = {Veldanda, A. K. and Zhang, S.-X. and Das, A. and Chakraborty, S. and Rawls, S. and Sahu, S. and Naphade, M.},
	year = {2024},
}

@article{ren_general_2025,
	title = {A general framework to enhance fine-tuning-based {LLM} unlearning},
	journal = {Findings of the Association for Computational Linguistics (ACL)},
	author = {Ren, J. and Dai, Z. and Tang, X. and Liu, H. and Zeng, J. and Li, Z. and Goutam, R. and Wang, S. and Xing, Y. and He, Q. and Liu, H.},
	year = {2025},
	pages = {18464--18476},
}

@incollection{heng_selective_2023,
	title = {Selective amnesia: a continual learning approach to forgetting in deep generative models},
	volume = {36},
	booktitle = {Advances in {Neural} {Information} {Processing} {Systems} ({NeurIPS})},
	author = {Heng, A. and Soh, H.},
	year = {2023},
	pages = {17170--17194},
}

@inproceedings{ho_classifier-free_2021,
	title = {Classifier-free diffusion guidance},
	booktitle = {{NeurIPS} {Workshop} on {Deep} {Generative} {Models} and {Downstream} {Applications} ({DGMs} and {Applications})},
	author = {Ho, J. and Salimans, T.},
	year = {2021},
}

@inproceedings{golatkar_eternal_2020,
	title = {Eternal sunshine of the spotless net: selective forgetting in deep networks},
	booktitle = {Proc. of the {IEEE} {Conf}. on {Computer} {Vision} and {Pattern} {Recognition} ({CVPR})},
	author = {Golatkar, A. and Achille, A. and Soatto, S.},
	year = {2020},
	pages = {9301--9309},
}

@inproceedings{gandikota_erasing_2023,
	title = {Erasing concepts from diffusion models},
	booktitle = {Proc. of the {IEEE}/{CVF} {Int}. {Conf}. on {Computer} {Vision} ({ICCV})},
	author = {Gandikota, R. and Materzyńska, J. and Fiotto-Kaufman, J. and Bau, D.},
	year = {2023},
	pages = {2426--2436},
}

@incollection{zhang_defensive_2024,
	title = {Defensive unlearning with adversarial training for robust concept erasure in diffusion models},
	volume = {37},
	booktitle = {Advances in {Neural} {Information} {Processing} {Systems} ({NeurIPS})},
	author = {Zhang, Y. and Chen, X. and Jia, J. and Zhang, Y. and Fan, C. and Liu, J. and Hong, M. and Ding, K. and Liu, S.},
	year = {2024},
	pages = {36748--36776},
}

@article{shi_retrack_2025,
	title = {{ReTrack}: data unlearning in diffusion models through redirecting the denoising trajectory},
	journal = {ArXiv: 2509.13007},
	author = {Shi, Q. and Jin, C. and Zhang, J. and Gu, Y.},
	year = {2025},
}

@article{epifano_revisiting_2023,
	title = {Revisiting the fragility of influence functions},
	volume = {162},
	journal = {Neural Networks},
	author = {Epifano, J. R. and Ramachandran, R. P. and Masino, A. J. and Rasool, G.},
	year = {2023},
	pages = {581--588},
}

@incollection{hu_most_2024,
	series = {37},
	title = {Most influential subset selection: challenges, promises, and beyond},
	booktitle = {Advances in {Neural} {Information} {Processing} {Systems} ({NeurIPS})},
	author = {Hu, Y. and Hu, P. and Zhao, H. and Ma, J. W.},
	year = {2024},
	pages = {119778--119810},
}

@article{k_revisiting_2021,
	title = {Revisiting methods for finding influential examples},
	journal = {ArXiv: 2111.04683},
	author = {K, K. and Søgaard, A.},
	year = {2021},
}

@book{pearl_causality_2013,
	edition = {2nd},
	title = {Causality},
	publisher = {Cambridge University Press},
	author = {Pearl, J.},
	year = {2013},
}

@article{morreale_attribution-by-design_2025,
	title = {Attribution-by-design: ensuring inference-time provenance in generative music systems},
	journal = {ArXiv: 2510.08062},
	author = {Morreale, F. and Hutiri, W. and Serrà, J. and Xiang, A. and Mitsufuji, Y.},
	year = {2025},
}

@article{wang_image_2004,
	title = {Image quality assessment: from error visibility to structural similarity},
	volume = {13},
	number = {4},
	journal = {IEEE Transactions on Image Processing},
	author = {Wang, Z. and Bovik, A. C. and Sheikh, H. R. and Simoncelli, E. P.},
	year = {2004},
	pages = {600--612},
}

@inproceedings{radford_learning_2021,
	title = {Learning transferable visual models from natural language supervision},
	booktitle = {Proc. of the {Int}. {Conf}. on {Machine} {Learning} ({ICML})},
	author = {Radford, A. and Kim, J. W. and Hallacy, C. and Ramesh, A. and Goh, G. and Agarwal, S. and Sastry, G. and Askell, A. and Mishkin, P. and Clark, J. and Krueger, G. and Sutskever, I.},
	year = {2021},
	pages = {8748--8763},
}

@article{oquab_dinov2_2023,
	title = {{DINOv2}: learning robust visual features without supervision},
	journal = {Transactions on Machine Learning Research},
	author = {Oquab, M and Darcet, T. and Moutakanni, T. and Vo, H. Y. and Szafraniec, M. and Khalidov, V. and Fernandez, P. and Haziza, D. and Massa, F. and El-Nouby, A. and Assran, M. and Ballas, N. and Galuba, W. and Howes, R. and Huang, P.-Y. and Li, S.-W. and Misra, I. and Rabbat, M. and Sharma, V. and Synnaeve, G. and Xu, H. and Jegou, H. and Mairal, J. and Labatut, P. and Joulin, A. and Bojanowski, P.},
	year = {2023},
}

@article{mason_areas_2002,
	title = {Areas beneath the relative operating characteristics ({ROC}) and relative operating levels ({ROL}) curves: statistical significance and interpretation},
	volume = {128},
	number = {584},
	journal = {Quarterly Journal of the Royal Meteorological Society},
	author = {Mason, S. J. and Graham, N. E.},
	year = {2002},
	pages = {2145--2166},
}

@article{krizhevsky_learning_2009,
	title = {Learning multiple layers of features from tiny images},
	journal = {Technical Report},
	author = {Krizhevsky, A.},
	year = {2009},
}

@article{liao_artbench_2022,
	title = {The {ArtBench} dataset: benchmarking generative models with artworks},
	journal = {ArXiv: 2206.11404},
	author = {Liao, P. and Li, X. and Liu, X. and Keutzer, K.},
	year = {2022},
}

@inproceedings{fang_captions_2015,
	title = {From captions to visual concepts and back},
	booktitle = {Proc. of the {IEEE} {Conf}. on {Computer} {Vision} and {Pattern} {Recognition} ({CVPR})},
	author = {Fang, H. and Gupta, S. and Iandola, F. and Srivastava, R. K. and Deng, L. and Dollár, P. and Gao, J. and He, X. and Mitchell, M. and Platt, J. C. and Zitnick, C. L. and Zweig, G.},
	year = {2015},
	pages = {1473--1482},
}

@inproceedings{peebles_scalable_2023,
	title = {Scalable diffusion models with transformers},
	booktitle = {Proc. of the {IEEE} {Int}. {Conf}. on {Computer} {Vision} ({ICCV})},
	author = {Peebles, W. and Xie, S.},
	year = {2023},
	pages = {4195--4205},
}

@incollection{paszke_pytorch_2019,
	series = {32},
	title = {{PyTorch}: an imperative style, high-performance deep learning library},
	booktitle = {Advances in {Neural} {Information} {Processing} {Systems} ({NeurIPS})},
	author = {Paszke, A. and Gross, S. and Massa, F. and Lerer, A. and Bradbury, J. and Chanan, G. and Killeen, T. and Lin, Z. and Gimelshein, N. and Antiga, L. and Desmaison, A. and Köpf, A. and Yang, E. and DeVito, Z. and Raison, M. and Tejani, A. and Chilamkurthy, S. and Steiner, B. and Fang, L. and Bai, J. and Chintala, S.},
	year = {2019},
}

@inproceedings{loshchilov_decoupled_2019,
	title = {Decoupled weight decay regularization},
	booktitle = {Proc. of the {Int}. {Conf}. on {Learning} {Representations} ({ICLR})},
	author = {Loshchilov, I. and Hutter, F.},
	year = {2019},
}

@article{bradley_use_1997,
	title = {The use of the area under the {ROC} curve in the evaluation of machine learning algorithms},
	volume = {30},
	number = {7},
	journal = {Pattern Recognition},
	author = {Bradley, A. P.},
	year = {1997},
	pages = {1145--1159},
}

@incollection{ho_denoising_2020,
	title = {Denoising diffusion probabilistic models},
	booktitle = {Advances in {Neural} {Information} {Processing} {Systems} ({NeurIPS})},
	author = {Ho, J. and Jain, A. and Abbeel, P.},
	year = {2020},
	pages = {6840--6851},
}

@inproceedings{wu_erasing_2025,
	title = {Erasing undesirable influence in diffusion models},
	booktitle = {Proc. of the {IEEE} {Conf}. on {Computer} {Vision} and {Pattern} {Recognition} ({CVPR})},
	author = {Wu, J. and Le, T. and Hayat, M. and Harandi, M.},
	year = {2025},
	pages = {28263--28273},
}

@inproceedings{zhang_unreasonable_2018,
	title = {The unreasonable effectiveness of deep features as a perceptual metric},
	booktitle = {Proc. of the {IEEE} {Conf}. on {Computer} {Vision} and {Pattern} {Recognition} ({CVPR})},
	author = {Zhang, R. and Isola, P. and Efros, A. A. and Shechtman, E. and Wang, O.},
	year = {2018},
	pages = {586--595},
}

@inproceedings{pizzi_self-supervised_2022,
	title = {A self-supervised descriptor for image copy detection},
	booktitle = {Proc. of the {IEEE} {Conf}. on {Computer} {Vision} and {Pattern} {Recognition} ({CVPR})},
	author = {Pizzi, E. and Roy, S. D. and Ravindra, S. N. and Goyal, P. and Douze, M.},
	year = {2022},
	pages = {14532--14542},

}

@inproceedings{deng_computational_2024,
	title = {Computational copyright: towards a royalty model for music generative {AI}},
	booktitle = {{ICLR} {Workshop} on {Navigating} and {Addressing} {Problems} for {Foundation} {Models} ({DPFM})},
	author = {Deng, J. and Ma, J.},
	year = {2024},
}

@inproceedings{kim_generation_2025,
	title = {From generation to attribution: music {AI} agent architectures for the post-streaming era},
	booktitle = {Proc. of the {AI} for {Music} {Workshop} at {NeurIPS25} ({AI4Music})},
	author = {Kim, W. and Wi, H. and Park, S. and Kim, T. and Keum, S. and Kim, K. and Kim, T. and Jung, J. and Kim, T. and Guerrero, G. and Le Goff, M. and Po, J. and Moon, D. and Nam, J. and Lee, J.},
	year = {2025},
}

@article{radovanovic_hubs_2010,
	title = {Hubs in space: popular nearest neighbors in high-dimensional data},
	volume = {11},
	journal = {Journal of Machine Learning Research},
	author = {Radovanović, M. and Nanopoulos, A. and Ivanović, M.},
	year = {2010},
	pages = {2487--2531},
}

@inproceedings{darcet_vision_2023,
	title = {Vision transformers need registers},
	booktitle = {Proc. of the {Int}. {Conf}. on {Learning} {Representations} ({ICLR})},
	author = {Darcet, T. and Oquab, M. and Mairal, J. and Bojanowski, P.},
	year = {2023},
}
}


\appendix

\section{Supplementary Methodology}
\label{sec:app_method}

\subsection{Pseudo-Code}
\label{sec:app_method_code}

Below we provide a Python-style pseudo-code of MUCS. Variables follow the notation in the main text (Sec.~\ref{sec:MUCS}). The full code is released at \link{}.

\begin{center}
\begin{varwidth}{0.9\linewidth}
{\small
\begin{Verbatim}[frame=lines,framesep=2mm,numbers=left,commandchars=\\\{\}]
\tcb{def} \tcy{MUCS}(x_hat, c_hat, M_1, L, Z, lambda=0.2):
    \tcg{# --- Constants ---}
    b = 100                                        \tcg{# Batch size}
    num_null = 20                                  \tcg{# To estimate L_null}
    epsilon = 0.001                                \tcg{# Prevent division by 0}
    \tcg{# --- Estimate L_null ---}
    M_0 = \tcy{get_copy}(M_1)
    M_0 = \tcy{random_init}(M_0)                         \tcg{# Randomly initialize}
    L_null = 0
    \tcb{for} _ \tcb{in} \tcy{range}(num_null):
        x, c = \tcy{sample_from}(Z, b)                   \tcg{# Get x,c in batches (b)}
        \tcg{# Get model predictions}
        sigma, n = \tcy{get_noise}(b)                    \tcg{# Original noise setup}
        x_pred = M_0(x, c, sigma, n)
        \tcg{# Get losses}
        L_est = L(x_pred, x)                       \tcg{# Original loss setup}
        \tcg{# Accumulate}
        L_null += \tcy{sum}(L_est)                       \tcg{# Sum over batch dim}
    L_null /= b * num_null
    \tcg{# --- Unlearn x_hat ---}
    M_2 = \tcy{get_copy}(M_1)
    x_hat, c_hat = \tcy{repeat}(x_hat, c_hat, b)         \tcg{# Repeat for batch size}
    \tcb{while} True:
        x, c = \tcy{sample_from}(Z, b)                   \tcg{# Get x,c in batches (b)}
        \tcg{# Get model predictions}
        sigma, n = \tcy{get_noise}(b)                    \tcg{# Original noise setup}
        x_pred = M_2(x, c, sigma, n)            
        sigma, n = \tcy{get_noise}(b)
        x_hat_pred = M_2(x_hat, c_hat, sigma, n)
        \tcg{# Get losses}
        L_ft = L(x_pred, x)                        \tcg{# Original loss setup}
        L_ft = \tcy{mean}(L_ft)                          \tcg{# Average over batch dim}
        L_ga = \tcy{min}(L(x_hat_pred, x_hat), L_null)
        L_ga = \tcy{mean}(L_ga)
        L_mucs = L_ft - lambda * L_ga
        \tcg{# Optimize}
        M_2 = \tcy{optimization_step}(M_2, L_mucs)       \tcg{# Original optimizer}
        \tcg{# Finished?}
        \tcb{if} L_ga >= 0.95 * L_null:
            \tcb{break}
    \tcg{# --- Compute attribution scores ---}
    _, n = \tcy{get_noise}(b)                            \tcg{# Original noise setup}
    sigma, _ = \tcy{get_noise_gen_sched}(b)              \tcg{# Generation schedule}
    A_mucs = \tcy{empty_array}(\tcy{len}(Z))
    \tcb{for} i \tcb{in} \tcy{range}(\tcy{len}(Z)):
        x, c = \tcy{repeat}(Z[i], b)                     \tcg{# Repeat for num. noises}
        \tcg{# Get model predictions}
        x_2_pred = M_2(x, c, sigma, n)      
        x_1_pred = M_1(x, c, sigma, n)
        \tcg{# Get losses}
        L_2 = L(x_2_pred, x)                       \tcg{# Original loss setup}
        L_1 = L(x_1_pred, x)
        \tcg{# Obtain normalized skew}
        A_mucs[i] = (L_2 - L_1) / (\tcy{abs}(L_2) + \tcy{abs}(L_1) + epsilon)
        A_mucs[i] = \tcy{mean}(A_mucs[i])                \tcg{# Average over noises}
    \tcg{# --- Done! ---}
    \tcb{return} A_mucs
\end{Verbatim}
}
\end{varwidth}
\end{center}

\subsection{Measuring Attribution Performance}
\label{sec:app_method_eval}

The evaluation of TDA algorithms for diffusion models is a challenging task~\cite{deng_survey_2025}. Some popular options are perceptual assessments, the linear datamodeling score, or indirect evaluation on related but different tasks. Perceptual assessments do not scale and, in general, perception might not correlate with model's behavior~\cite{morreale_attribution-by-design_2025, pearl_causality_2013}. The linear datamodeling score presents a large number of issues discouraging its use~\cite{epifano_revisiting_2023, hu_most_2024}. Indirect evaluation is not addressing attribution itself, and is often performed on synthetic/toy problems which may not represent a real-world situation~\cite{deng_survey_2025, k_revisiting_2021}. In this paper, we evaluate TDA performance using a leave-$k$-out counterfactual setup~\cite{deng_survey_2025, wang_data_2024}, in which one tests the ability of a model to generate a previously generated image after retraining from scratch without including its influential items in the training set. Albeit time-consuming, we believe it is the only reliable way to assess the performance of TDA algorithms in diffusion models (we introduce a slight modification to parallelize the computational cost of this evaluation, see below). 

Given a training dataset $\se{Z}$ and a diffusion model $F_1$ trained with it, we generate $\hat{\ve{z}}$ using some random seed $u$ and a conditioning $\ve{c}$ from the test set. Next, an attribution algorithm $A$ provides attribution scores $a_i$ for each $\ve{z}_i\in\se{Z}$. These scores are then ranked from highest to lowest and the instances with top-$k$ scores are removed from $\se{Z}$, creating $\se{Z}'$. We retrain from scratch the same model $F$ on $\se{Z}'$, producing $F_1'$, and generate $\hat{\ve{z}}'$ using the same $u$ and $\ve{c}$ used for $\hat{\ve{z}}$. Because of the seed-consistency property of diffusion models~\cite{georgiev_journey_2023, wang_data_2024}, $\hat{\ve{z}}$ and $\hat{\ve{z}}'$ will show a high similarity when $A$ fails to identify/attribute influential items (e.g.,~when $A$ just randomly selects training items; examples are available in~\cite{georgiev_journey_2023, wang_data_2024} and in Fig.~\ref{fig:image_pairs_random}). Only when relevant influential samples have been removed from $\se{Z}_\tx{train}$, the model $F_1'$ will generate dissimilar\footnote{Intuitively, removing influential items here corresponds to removing nearest neighbors in the distribution `seen' by the model, hence unpopulating the region where the generated sample lies. When that happens, the newly trained model is nudged to generate a different item given the same seed (and conditioning).} instances $\hat{\ve{z}}'$ under the same $u$ and $\ve{c}$ (examples are available in~\cite{georgiev_journey_2023, wang_data_2024} and in Figs.~\ref{fig:imchange_cifar}--\ref{fig:imchange_coco}). To avoid retraining a model for each $\hat{\ve{z}}$, we can speed up the previous procedure by conducting $m$ generations in parallel and removing $mk$ items from $\se{Z}$. 

By comparing $\hat{\ve{z}}$ and $\hat{\ve{z}}'$ generations, we can have a nice picture of the performance of an attribution algorithm: a low similarity highlights the degree to which the removed items impede the generation of $\hat{\ve{z}}$. 
To objectively measure image similarity, and to avoid any bias towards a particular metric, we rely on four different but complementary measures: the structural similarity index measure (SSIM)~\cite{wang_image_2004}, cosine similarity in the SSCD copy-detection space (SSCD)~\cite{pizzi_self-supervised_2022}, the learned perceptual image patch similarity (LPIPS) measure~\cite{zhang_unreasonable_2018}, and cosine similarity in the semantic CLIP space (CLIP)~\cite{radford_learning_2021}. Fig.~\ref{fig:example_distribs} shows examples of the distance distributions we get for both random attribution and a considered approach. In the figure, we see that SSIM values concentrate between 0.8 and 1 for the case of random removal, and that MUCS can shift the distance distribution to values between 0.4 and 0.9. 

Our final assessment is based on comparing the similarity distribution of the considered algorithms $A$ with the one obtained by random removal $A_\tx{rand}$ (our reference similarity distribution). In particular, we are interested in a low overlap of the two distributions, indicating that removing training instances according to $A$ had a meaningful effect. To objectively quantify the overlap between two distributions, we use the area under the ROC curve (AUC) measure, which is a common evaluation measure employed for that purpose~\cite{bradley_use_1997}, and which has the nice property of being proportional to the U statistic of the Mann-Whitney U test for measuring whether two samples correspond to the same distribution~\cite{mason_areas_2002}. This way, a good attribution algorithm should achieve an AUC close to 1, corresponding to a low overlap, indicating that attributed items' removal is affecting generation. 

Our final setup involves removing $k=2$\% of the training data for $m=20$ generations, and running it 6~times using different network and seed initializations. Hence, AUC calculations are made on pairs of 120~generated items (random and considered approach). Notice that removing $km$ items corresponds to removing a maximum of 40\% of the training data (there can be overlaps between the $m$ items). Notice also that, ideally, the value of $k$ should depend on each dataset and generated item. Here, it was empirically set such that AUCs were significantly larger than random, and such that less than half of the training data was removed. 

\subsection{Model, Training, and Generation}
\label{sec:app_method_model}

\topic{Architecture}
As mentioned in the main text, we use a DiT/B architecture~\cite{peebles_scalable_2023}, which consists of 12~Transformer blocks of 768~channels, 12~heads, an MLP ratio of 4, and a dropout of 0.1. To adapt different image resolutions to this architecture, we use two residual convolutional blocks with a number of channels and strides that depend on the dataset (Table~\ref{tab:training_confs}). These blocks use a kernel of 5 with same-size padding, instance norm, and the SiLU activation (the decoder blocks use transposed convolutions). After image encoding with this structure, and together with positional encoding, we add 32~extra tokens to be used as memory registers~\cite{darcet_vision_2023}. 

The conditioning, as well as the $\sigma$ embeddings, are processed by a two-layer MLP with SiLU activations and a dropout of 0.1. The number of channels of the conditioning module varies depending on the density of the information contained in the conditioning vector (compare ArtBench10 and COCO in Table~\ref{tab:training_confs}). For conditional models, we train with a condition drop of 10\% to enable classifier-free guidance (CFG) at sampling. Note that, as mentioned in the main text, we consider CIFAR10 as unconditional, ArtBench10 is conditioned on the class identifier, and COCO is conditioned on CLIP-Text embeddings. All network weights are initialized following PyTorch's defaults~\cite{paszke_pytorch_2019}.

\topic{Diffusion Variant and Generation}
We consider the EDM diffusion variant~\cite{karras_elucidating_2022}, using $P_\tx{mean}=-1.2$, $P_\tx{std}=1.2$, and $\sigma_\tx{data}=0.5$ for all datasets (data is normalized between $-$1 and 1). As illustrated in the main text, the loss includes the characteristic EDM weighting term (Sec.~\ref{sec:MUCS_overview}). For generation, we use $\sigma_\tx{min}=2\cdot10^{-3}$, $\sigma_\tx{max}=80$, $\rho=7$, and 32 steps of the 2$^\tx{nd}$ order Heun sampler (we use deterministic sampling). The CFG weight depends on the dataset (see Table~\ref{tab:training_confs}).

\topic{Training}
The training of our diffusion models consists of hundreds of thousands of weight updates using AdamW~\cite{loshchilov_decoupled_2019} with a variable learning rate and a weight decay of 0.01 (details about number of updates and nominal learning rate for each dataset can be found in Table~\ref{tab:training_confs}). We perform a linear learning rate warmup of 10\,k steps and maintain it constant for the rest of the updates. Our batch size is 128, and we use an exponential moving average of the model's weights~\cite{karras_elucidating_2022} with a momentum of 0.999.

\begin{table}[h]
  \vspace{0.3cm}
  \caption{Configuration differences for each dataset. The rest of the parameters follow the canonical DiT/B and EDM setups (see text).}
  \label{tab:training_confs}
  \centering
  \begin{tabular}{lccc}
    \toprule
    \textbf{Parameter} & \multicolumn{3}{c}{\textbf{Dataset}} \\
    \cmidrule(r){2-4}
              					& CIFAR10		& ArtBench10	& COCO \\
    \midrule
    Image resolution			& 32$\times$32  & 64$\times$64  & 64$\times$64 \\
	Encoder/decoder channels	& [4,4]			& [16,16]		& [16,16] \\
	Encoder/decoder strides		& [1,1]			& [1,2]			& [1,2] \\
    Conditioning				& None  		& Class ID		& CLIP-Text \\
	Conditioning block channels & 256			& 256			& 768 \\
    Total learnable parameters 	& 100\,M		& 100\,M	    & 146\,M \\
    Learning rate               & 10$^{-4}$     & 2$\cdot$10$^{-4}$ & 2$\cdot$10$^{-4}$ \\
    Pre-training updates        & 200\,k        & 250\,k        & 300\,k \\
	CFG weight     				& N/A			& 2.0			& 3.0 \\
    \bottomrule
  \end{tabular}
  \vspace{0.3cm}
\end{table}

\subsection{Rank-based TDA Ensemble}
\label{sec:rankensemble}

To produce a TDA ensemble, we take the scores $\ve{a}$ of every approach and transform them into ranks $\ve{r}$ following a descending order. We then weight $\ve{r}$ and sum across the approaches that form the ensemble. Weights are loosely assigned based on the average performances reported in Table~\ref{tab:comparison} in the main paper. Following that, we set the weight of MUCS to 10, DAS to 7, D-TRAK to 6, DINO to 5, CLIP to 5, Condition to 3, Forward-INF to 2, and AbU to 0. 
After summing the weighted rankings, we take the inverse as a new attribution score (descending order) and evaluate. In preliminary experiments, we employed a simple 0--1 normalization instead of ranks, but we obtained slightly worse results. 

We acknowledge that such an ensembling procedure can be further improved, but leave this task for future works. In this paper, we just aim to perform an initial exploration of the suitability of ensembling different attribution methods, and we believe the considered approach is sufficient for that purpose. To the best of our knowledge, this is the first time that such an ensembling approach has been considered for TDA. As mentioned in the main paper, our contribution is to show that it can provide important boosts in some cases, but that one should be careful in others (Sec.~\ref{sec:results}).


\section{Additional Results}
\label{sec:app_results}

\subsection{Comparison with Existing Approaches}
\label{sec:app_results_compare}

To complement the results in the main paper, we here show the average normalized similarity difference between $A_\tx{rand}$ and each considered approach (Table~\ref{tab:comparison_mean}). We calculate the average normalized similarity difference as
\begin{equation*}
\frac{100}{mk} \sum_{i=1,\dots mk} \left( \frac{s_i}{\tx{median}(\se{S})} -1\right) ,
\end{equation*}
where $s_i=S(\hat{\ve{x}}_i,\hat{\ve{x}}'_i)$ denotes the corresponding image similarity metric and $\se{S}$ denotes the set of all $A_\tx{rand}$ vs.\ $A$ similarities (see Sec.~\ref{sec:eval} and Appendix~\ref{sec:app_method_eval}). In the case of LPIPS, we convert its loss $L_\text{LPIPS}$ into a similarity measure $s$ by simply using $s=\max(-1,1-L_\text{LPIPS})$.

\begin{figure}[h]
\centering
\includegraphics[width=0.8\linewidth]{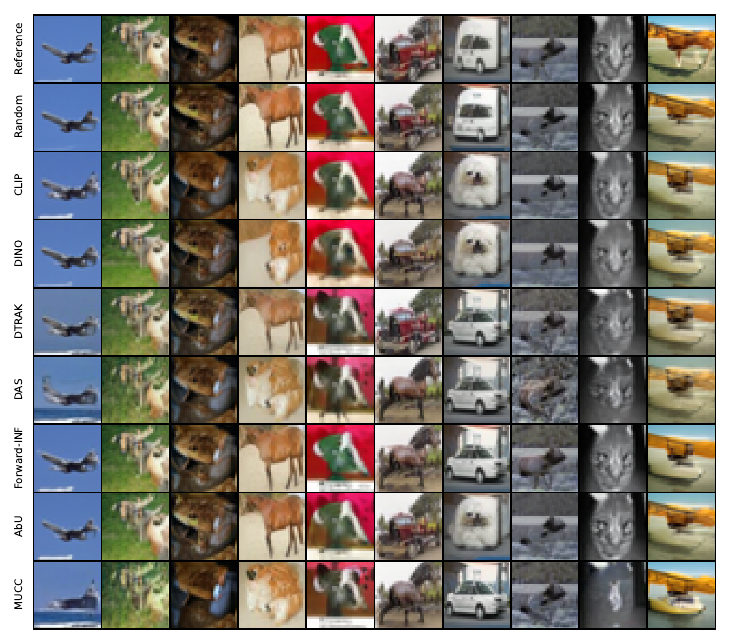}
\caption{Reference (first row) and regenerated (other rows) images for the considered approaches on CIFAR10.}
\label{fig:imchange_cifar}
\end{figure}

\begin{figure}[h]
\centering
\includegraphics[width=0.8\linewidth]{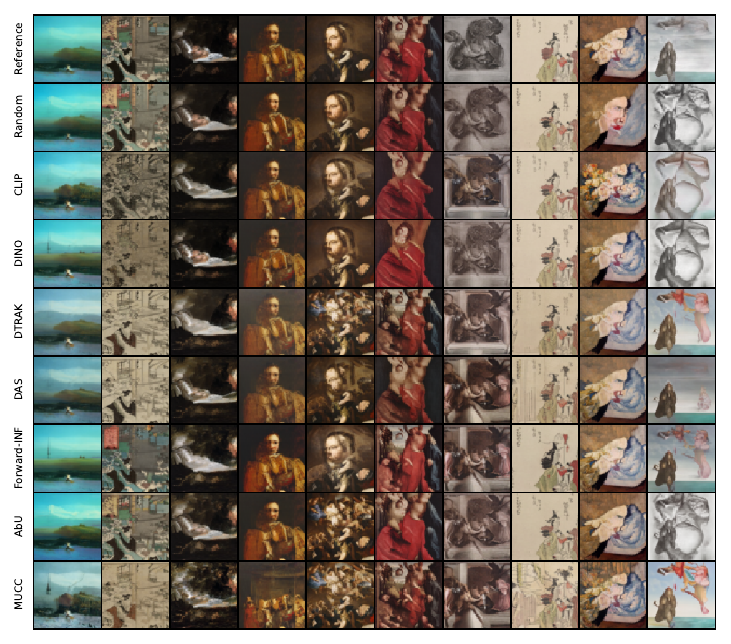}
\caption{Reference (first row) and regenerated (other rows) images for the considered approaches on ArtBench10.}
\label{fig:imchange_artbench10}
\end{figure}

\begin{figure}[h]
\centering
\includegraphics[width=0.8\linewidth]{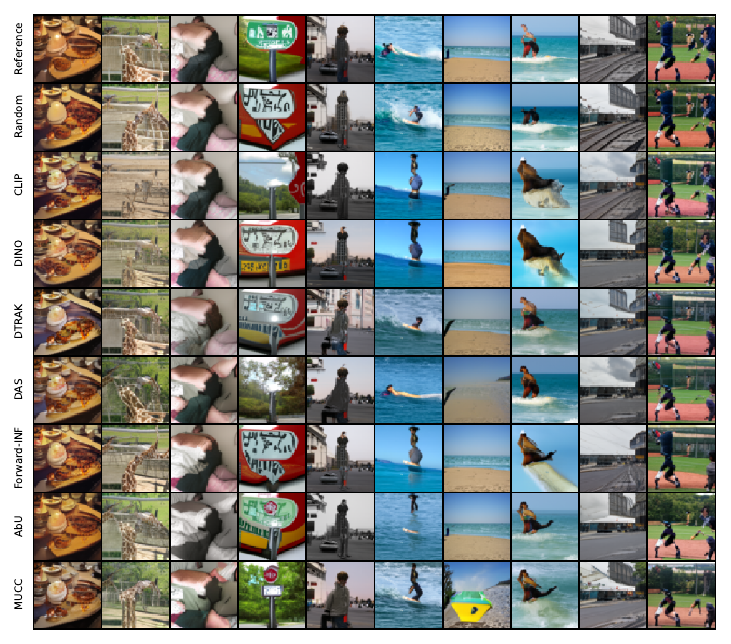}
\caption{Reference (first row) and regenerated (other rows) images for the considered approaches on COCO.}
\label{fig:imchange_coco}
\end{figure}

Apart from assessing distribution overlaps, we also report the amortized running time of the considered TDA approaches (Table~\ref{tab:compute_time}). Finally, our results also include a sample of the images generated for each dataset, before and after attribution-based removal, comparing reference (generated with $F_1$) and regenerated (generated with $F_2$) images (Figs.~\ref{fig:imchange_cifar}--\ref{fig:imchange_coco}).

\vfill

\begin{table}[h]
  \vspace{0.3cm}
  \caption{Comparison with existing approaches on CIFAR10 (top), ArtBench10 (middle), and COCO (bottom). Approaches can be categorized into model-agnostic (MA), influence function-based (IF), and unlearning-based (U). We report average similarity differences (\%, the lower the better) for the considered measures, plus their average ($\mu$; last column). The $\pm$ signs indicate 95\% confidence intervals.}
  \label{tab:comparison_mean}
  \centering
  \resizebox{\columnwidth}{!}{%
  \renewcommand{\arraystretch}{1.1}
  \begin{tabular}{llcrrrrr}
    \toprule
     & \textbf{Approach} & \textbf{Categ.} 	& \textbf{SSIM}~~~ & \textbf{SSCD}~~~ & \textbf{LPIPS}~~ & \textbf{CLIP}~~~ & $\bm\mu$~~~~~~ \\
    \midrule
    \multirow{8}{*}{\rotatebox[origin=c]{90}{CIFAR10}} 
    & Condition 							& MA & $-$2.5$\pm$0.9 & $-$4.5$\pm$1.7 & $-$2.6$\pm$1.4 & $-$3.6$\pm$1.9 & $-$3.3$\pm$1.5 \\
    & CLIP	 								& MA & $-$4.8$\pm$1.4 & $-$8.9$\pm$2.2 & $-$6.1$\pm$1.7 & $-$7.8$\pm$2.3 & $-$6.9$\pm$1.9 \\
    & DINO	 					            & MA & $-$4.5$\pm$1.4 & $-$7.5$\pm$2.1 & $-$5.5$\pm$1.7 & $-$6.4$\pm$2.3 & $-$6.0$\pm$1.9 \\
    & D-TRAK~\cite{zheng_intriguing_2024}	& IF & $-$8.1$\pm$1.9 & $-$11.0$\pm$2.4 & $-$9.7$\pm$2.1 & $-$7.3$\pm$2.3 & $-$9.0$\pm$2.2 \\
    & DAS~\cite{lin_diffusion_2024} 		& IF & $-$11.2$\pm$2.1 & $-$15.5$\pm$2.6 & $-$13.1$\pm$2.2 & $-$11.7$\pm$2.4 & $-$12.9$\pm$2.3 \\
    & Forward-INF~\cite{ko_mirrored_2024} 	& U  & $-$3.6$\pm$1.5 & $-$3.6$\pm$2.0 & $-$2.9$\pm$1.7 & $-$2.5$\pm$1.9 & $-$3.1$\pm$1.8 \\
    & AbU~\cite{wang_data_2024} 			& U  & $-$2.8$\pm$1.2 & $-$2.1$\pm$1.7 & $-$1.3$\pm$1.4 & $-$0.8$\pm$1.7 & $-$1.7$\pm$1.5 \\
    & MUCS (proposed)    					& U  & $-$13.5$\pm$2.1 & $-$20.8$\pm$2.3 & $-$17.9$\pm$1.8 & $-$18.1$\pm$2.1 & $-$17.6$\pm$2.1 \\
    \midrule
    \multirow{8}{*}{\rotatebox[origin=c]{90}{ArtBench10}} 
    & Condition 							& MA & $-$3.0$\pm$1.3 & $-$6.7$\pm$2.9 & $-$4.0$\pm$1.8 & $-$2.9$\pm$1.3 & $-$4.2$\pm$1.8 \\
    & CLIP	 								& MA & $-$5.2$\pm$1.6 & $-$12.5$\pm$3.3 & $-$7.6$\pm$2.0 & $-$7.3$\pm$1.8 & $-$8.1$\pm$2.2 \\
    & DINO	 					            & MA & $-$3.6$\pm$1.5 & $-$8.5$\pm$3.1 & $-$4.6$\pm$1.8 & $-$4.1$\pm$1.5 & $-$5.2$\pm$2.0 \\
    & D-TRAK~\cite{zheng_intriguing_2024}	& IF & $-$10.5$\pm$2.0 & $-$20.9$\pm$3.9 & $-$13.4$\pm$2.4 & $-$9.4$\pm$2.1 & $-$13.6$\pm$2.6 \\
    & DAS~\cite{lin_diffusion_2024} 		& IF & $-$11.1$\pm$2.0 & $-$21.3$\pm$3.6 & $-$14.1$\pm$2.3 & $-$9.0$\pm$2.0 & $-$13.9$\pm$2.5 \\
    & Forward-INF~\cite{ko_mirrored_2024} 	& U  & $-$6.4$\pm$1.7 & $-$11.7$\pm$3.3 & $-$7.6$\pm$2.0 & $-$5.4$\pm$1.8 & $-$7.8$\pm$2.2 \\
    & AbU~\cite{wang_data_2024} 			& U  & $-$3.2$\pm$1.3 & $-$5.0$\pm$2.8 & $-$3.3$\pm$1.7 & $-$2.9$\pm$1.4 & $-$3.6$\pm$1.8 \\
    & MUCS (proposed)    					& U  & $-$12.8$\pm$1.9 & $-$25.4$\pm$3.2 & $-$17.0$\pm$2.1 & $-$11.9$\pm$1.7 & $-$16.8$\pm$2.2 \\
    \midrule
    \multirow{8}{*}{\rotatebox[origin=c]{90}{COCO}} 
    & Condition 							& MA & $-$7.4$\pm$2.3 & $-$12.8$\pm$4.3 & $-$5.5$\pm$2.2 & $-$5.4$\pm$1.8 & $-$7.8$\pm$2.7 \\
    & CLIP	 								& MA & $-$11.8$\pm$3.0 & $-$21.4$\pm$5.1 & $-$11.7$\pm$2.9 & $-$12.1$\pm$2.3 & $-$14.3$\pm$3.3 \\
    & DINO	 					            & MA & $-$11.9$\pm$2.9 & $-$21.5$\pm$4.9 & $-$11.0$\pm$2.8 & $-$10.8$\pm$2.0 & $-$13.8$\pm$3.2 \\
    & D-TRAK~\cite{zheng_intriguing_2024}	& IF & $-$13.3$\pm$2.8 & $-$18.1$\pm$4.3 & $-$12.4$\pm$2.7 & $-$6.8$\pm$1.7 & $-$12.6$\pm$2.9 \\
    & DAS~\cite{lin_diffusion_2024} 		& IF & $-$13.9$\pm$3.0 & $-$20.3$\pm$4.7 & $-$12.6$\pm$2.9 & $-$7.9$\pm$2.0 & $-$13.7$\pm$3.2 \\
    & Forward-INF~\cite{ko_mirrored_2024} 	& U  & $-$8.2$\pm$2.6 & $-$11.6$\pm$4.4 & $-$6.4$\pm$2.6 & $-$5.6$\pm$2.0 & $-$8.0$\pm$2.9 \\
    & AbU~\cite{wang_data_2024} 			& U  & $-$5.4$\pm$2.4 & $-$3.9$\pm$3.8 & $-$3.0$\pm$2.1 & $-$2.8$\pm$1.6 & $-$3.8$\pm$2.5 \\
    & MUCS (proposed)    					& U  & $-$15.0$\pm$2.9 & $-$24.9$\pm$4.8 & $-$16.2$\pm$3.0 & $-$8.9$\pm$2.0 & $-$16.2$\pm$3.2 \\
   \bottomrule
  \end{tabular}
  }
  \vspace{0.3cm}
\end{table}
\vfill

\begin{table}[h]
  \vspace{0.3cm}
  \caption{Amortized time (in minutes) to compute attribution for a single generated item, using a single NVIDIA H100 GPU. Time depends on the size of the training dataset and, in this table, includes the proportion of the time to initialize all operations (less than 0.1~minutes).}
  \label{tab:compute_time}
  \centering
  \resizebox{0.6\columnwidth}{!}{%
  \begin{tabular}{lrrr}
    \toprule
    \textbf{Approach} & \textbf{CIFAR10} & \textbf{ArtBench10} & \textbf{COCO} \\
    \midrule
    Random      & $<$0.1 & $<$0.1  & 0.1 \\
    Condition   & $<$0.1 & $<$0.1  & 0.2 \\
    CLIP        & 0.1    & 0.1     & 0.3 \\
    DINO        & $<$0.1 & $<$0.1  & 0.2 \\
    D-TRAK      & 57.4   & 36.0    & 101.1 \\
    DAS         & 66.7   & 40.8    & 113.9 \\
    Forward-INF & 34.3   & 36.1    & 82.8 \\
    AbU         & 36.4   & 38.1    & 87.9 \\
    MUCS        & 22.8   & 24.2    & 56.5 \\
    \bottomrule
  \end{tabular}
  }
  \vspace{0.3cm}
\end{table}




\FloatBarrier

\subsection{Overlap Between Generated Items}
\label{sec:app_results_overlap}

Table~\ref{tab:overlap_items} shows the average overlap between top-$k$ attributed instances. First, for $m$ generated items, we obtain one top-$k$ array of topmost influential training instances. Next, independently for every attribution approach, we compute the overlap (intersection) across items, which results in $m(m-1)/2$ combinations. Finally, we take the average across those combinations. 

We see that there is always an overlap (that is, overlap never significantly close to 0), and that this usually ranges between 2 and 7\% of the top-$k$ instances. The relatively low values for the latter indicate that the removed items are generally different for the $m$ generated instances. It also indicates that the attribution methods do not exploit any kind of massive generative model disruption (e.g.,~targeting only and always the same `hub'/structural training items for every generated instance).

\begin{table}[h]
  \vspace{0.3cm}
  \caption{Average top-$k$ attributed instances' overlap (\%) between generated items. The $\pm$ signs denote 95\% confidence intervals.}
  \label{tab:overlap_items}
  \centering
  \resizebox{0.6\columnwidth}{!}{%
  \begin{tabular}{lrrr}
    \toprule
    \textbf{Approach} & \textbf{CIFAR10} & \textbf{ArtBench10} & \textbf{COCO}~~~ \\
    \midrule
    Random      & 2.0 $\pm$ 0.0 & 2.0 $\pm$ 0.0~~ & 2.0 $\pm$ 0.0 \\
    Condition   & 2.0 $\pm$ 0.1 & 1.9 $\pm$ 0.1~~ & 2.5 $\pm$ 0.2 \\
    CLIP        & 5.9 $\pm$ 0.2 & 6.5 $\pm$ 0.2~~ & 4.1 $\pm$ 0.2 \\
    DINO        & 6.0 $\pm$ 0.2 & 6.1 $\pm$ 0.2~~ & 3.4 $\pm$ 0.2 \\
    D-TRAK      & 3.6 $\pm$ 0.2 & 2.6 $\pm$ 0.2~~ & 3.8 $\pm$ 0.2 \\
    DAS         & 3.9 $\pm$ 0.3 & 2.1 $\pm$ 0.2~~ & 2.3 $\pm$ 0.1 \\
    Forward-INF & 10.9 $\pm$ 0.4 & 2.9 $\pm$ 0.2~~ & 3.0 $\pm$ 0.2 \\
    AbU         & 32.0 $\pm$ 0.8 & 39.6 $\pm$ 0.8~~ & 25.9 $\pm$ 0.6 \\
    MUCS        & 2.7 $\pm$ 0.2 & 3.0 $\pm$ 0.2~~ & 4.2 $\pm$ 0.2 \\
    \bottomrule
  \end{tabular}
  }
  \vspace{0.3cm}
\end{table}

\subsection{Ensembling Approaches}
\label{sec:app_results_ensemble}

Continuing with the overlap idea, we can compute top-$k$ instance overlap between TDA approaches. Comparing pairwise, this tells us how different are the top-$k$ selected items across approaches (Table~\ref{tab:overlap_appr}). Interestingly, we see that the overlap is usually below 35\%, which suggests that an ensemble approach could benefit from these multiple criteria and obtain better performance. After running the aforementioned rank-based ensemble approach (Appendix~\ref{sec:rankensemble}), we obtain mixed results depending on the dataset (Table~\ref{tab:ensemble_extend}; see the comments in the main paper).

\begin{table}[h]
  \vspace{0.3cm}
  \caption{Average top-$k$ overlap (\%) between methods for ArtBench10, comparing each generated item and averaging (see text).}
  \label{tab:overlap_appr}
  \centering
  \resizebox{1\columnwidth}{!}{%
  \begin{tabular}{lrrrrrrrrr}
    \toprule
    \textbf{Approach}~~$\bm{\downarrow~\rightarrow}$ & Rand & Cond & CLIP & DINO & DTRK & DAS & F-INF & AbU & MUCS \\
    \midrule
    Random      & - & 2.0 & 2.0 & 2.0 & 2.0 & 2.0 & 2.0 & 2.0 & 2.0 \\
    Condition   & 2.0 & - & 8.1 & 6.3 & 13.4 & 16.2 & 17.4 & 4.2 & 10.3 \\
    CLIP        & 2.0 & 8.1 & - & 28.2 & 13.6 & 13.9 & 6.8 & 2.9 & 19.9 \\
    DINO        & 2.0 & 6.3 & 28.2 & - & 9.8 & 9.9 & 5.6 & 2.4 & 14.6 \\
    D-TRAK      & 2.0 & 13.4 & 13.6 & 9.8 & - & 60.7 & 12.2 & 6.9 & 36.0 \\
    DAS         & 2.0 & 16.2 & 13.9 & 9.9 & 60.7 & - & 13.3 & 5.6 & 32.5 \\
    Forward-INF & 2.0 & 17.4 & 6.8 & 5.6 & 12.2 & 13.3 & - & 8.0 & 8.5 \\
    AbU         & 2.0 & 4.2 & 2.9 & 2.4 & 6.9 & 5.6 & 8.0 & - & 6.3 \\
    MUCS        & 2.0 & 10.3 & 19.9 & 14.6 & 36.0 & 32.5 & 8.5 & 6.3 & - \\
    \bottomrule
  \end{tabular}
  }
  \vspace{0.3cm}
\end{table}

\begin{table}[h]
  \vspace{0.3cm}
  \caption{MUCS (R) and ensemble results (A--D) for CIFAR10 (top), ArtBench10 (middle), and COCO (bottom).}
  \label{tab:ensemble_extend}
  \centering
  \setlength{\tabcolsep}{6pt}
  \resizebox{\columnwidth}{!}{%
  \begin{tabular}{cccccccccccc}
    \toprule
    \textbf{\#} & \multicolumn{7}{c}{\textbf{Weight in the Ensemble}} & \multicolumn{4}{c}{\textbf{AUC}} \\
    \cmidrule{9-12}
     & MUCS & DAS & D-TRAK & DINO & CLIP & Cond & F-INF & \textbf{SSIM} & \textbf{SSCD} & \textbf{LPIPS} & \textbf{CLIP} \\
    \midrule
    R & 10 & 0 & 0 & 0 & 0 & 0 & 0 & 0.889 & 0.901 & 0.930 & 0.868 \\
    A & 10 & 7 & 6 & 5 & 5 & 3 & 2 & 0.825 & 0.843 & 0.864 & 0.803 \\
    B & 10 & 0 & 0 & 5 & 5 & 3 & 0 & 0.696 & 0.706 & 0.734 & 0.716 \\
    C &  0 & 7 & 6 & 5 & 5 & 3 & 2 & 0.775 & 0.788 & 0.805 & 0.761 \\
    D &  0 & 0 & 0 & 5 & 5 & 3 & 0 & 0.593 & 0.612 & 0.613 & 0.604 \\
    \midrule
    R & 10 & 0 & 0 & 0 & 0 & 0 & 0 & 0.850 & 0.840 & 0.859 & 0.823 \\
    A & 10 & 7 & 6 & 5 & 5 & 3 & 2 & 0.858 & 0.858 & 0.878 & 0.843 \\
    B & 10 & 0 & 0 & 5 & 5 & 3 & 0 & 0.780 & 0.816 & 0.821 & 0.835 \\
    C &  0 & 7 & 6 & 5 & 5 & 3 & 2 & 0.797 & 0.803 & 0.824 & 0.815 \\
    D &  0 & 0 & 0 & 5 & 5 & 3 & 0 & 0.682 & 0.712 & 0.706 & 0.736 \\
    \midrule
    R & 10 & 0 & 0 & 0 & 0 & 0 & 0 & 0.750 & 0.763 & 0.799 & 0.715 \\
    A & 10 & 7 & 6 & 5 & 5 & 3 & 2 & 0.784 & 0.803 & 0.822 & 0.795 \\
    B & 10 & 0 & 0 & 5 & 5 & 3 & 0 & 0.776 & 0.803 & 0.829 & 0.795 \\
    C &  0 & 7 & 6 & 5 & 5 & 3 & 2 & 0.743 & 0.763 & 0.782 & 0.758 \\
    D &  0 & 0 & 0 & 5 & 5 & 3 & 0 & 0.715 & 0.743 & 0.757 & 0.806 \\
    \bottomrule
  \end{tabular}
  }
  \vspace{0.3cm}
\end{table}


\section{Further Information}

\subsection{Licenses}
\label{sec:app_licenses}

As mentioned in the main text, we use the CIFAR10~\cite{krizhevsky_learning_2009}, ArtBench10~\cite{liao_artbench_2022}, and COCO~\cite{fang_captions_2015} datasets. CIFAR10 is under MIT license, ArtBench10 is under Fair Use license, COCO images are under Flickr Terms of Use license (various Creative Commons), and COCO captions is under the Creative Commons Attribution 4.0 license. As for pre-trained models, SSIM~\cite{wang_image_2004}, SSCD~\cite{pizzi_self-supervised_2022}, and CLIP~\cite{radford_learning_2021} are under MIT license, LPIPS~\cite{zhang_unreasonable_2018} is under BSD 2-Clause license, and DINOv2~\cite{oquab_dinov2_2023} is under Apache-2.0 license. We release our code under Apache-2.0 license.

\subsection{Compute Resources}
\label{sec:app_compute}

Training one of the considered diffusion models takes between 1 and 2~days, using a single NVIDIA H100 GPU with 80\,GB memory (depending if the dataset is small, like CIFAR10, or larger, like COCO). Computing attribution for 20~items takes a maximum of 1 day using the same resource (one full forward pass per training instance, repeated 20~times; see also Table~\ref{tab:compute_time}). The time to evaluate results is just seconds. Hence, the full pipeline (training $F_1$, computing $A$, training $F_2$, and evaluating) takes roughly 3 to 5~days.

\subsection{Limitations and Broader Impact}
\label{sec:app_limitations}


While MUCS demonstrates strong empirical performance, its theoretical guarantees remain limited, as is common in the TDA literature. Attribution reliability may still degrade with different, larger-scale datasets, or with highly redundant or overly sparse training data. Our evaluation is also restricted to diffusion models and image data, and generalization to other generative paradigms or domains is not guaranteed. Additionally, the absence of ground-truth attribution scores makes absolute performance assessment inherently approximate, and the computational overhead of performing a full forward pass for each training instance may limit applicability at very large scales or in real-world situations.

TDA has the potential to significantly improve the transparency and accountability of generative models. From a socio-technical perspective, reliable TDA could empower anyone with access to training data and model weights to audit the influence of specific data points on generated outputs, with direct implications for intellectual property, consent, and data governance frameworks. Improved unlearning capabilities could further support right-to-be-forgotten compliance under regulations such as GDPR. However, TDA tools could also be misused to reverse-engineer proprietary datasets or circumvent data protection measures, raising dual-use concerns. As generative AI becomes more pervasive, tools like MUCS could help mitigate harms from non-consensual data use, but their deployment should be accompanied by appropriate governance structures to avoid over-reliance on purely technical solutions.


\ifdefined\anon
\newpage
\section*{NeurIPS Paper Checklist}

\begin{enumerate}

\item {\bf Claims}
    \item[] Question: Do the main claims made in the abstract and introduction accurately reflect the paper's contributions and scope?
    \item[] Answer: \answerYes{} 
    \item[] Justification: Abstract and introduction contain a concise summary of the paper's contributions.

\item {\bf Limitations}
    \item[] Question: Does the paper discuss the limitations of the work performed by the authors?
    \item[] Answer: \answerYes{} 
    \item[] Justification: See Sec.~\ref{sec:results} and Appendix~\ref{sec:app_limitations}.

\item {\bf Theory assumptions and proofs}
    \item[] Question: For each theoretical result, does the paper provide the full set of assumptions and a complete (and correct) proof?
    \item[] Answer: \answerNA{} 
    \item[] Justification: \answerNA{}

    \item {\bf Experimental result reproducibility}
    \item[] Question: Does the paper fully disclose all the information needed to reproduce the main experimental results of the paper to the extent that it affects the main claims and/or conclusions of the paper (regardless of whether the code and data are provided or not)?
    \item[] Answer: \answerYes{} 
    \item[] Justification: Both the main document and the Appendix disclose all the information.

\item {\bf Open access to data and code}
    \item[] Question: Does the paper provide open access to the data and code, with sufficient instructions to faithfully reproduce the main experimental results, as described in supplemental material?
    \item[] Answer: \answerYes{} 
    \item[] Justification: All the data is publicly available, pseudo-code is provided in Appendix~\ref{sec:app_method_code}, and the full code will be released under Apache-2.0 license after the review phase.

\item {\bf Experimental setting/details}
    \item[] Question: Does the paper specify all the training and test details (e.g., data splits, hyperparameters, how they were chosen, type of optimizer) necessary to understand the results?
    \item[] Answer: \answerYes{} 
    \item[] Justification: See Sec.~\ref{sec:eval} and Appendix~\ref{sec:app_method}.

\item {\bf Experiment statistical significance}
    \item[] Question: Does the paper report error bars suitably and correctly defined or other appropriate information about the statistical significance of the experiments?
    \item[] Answer: \answerYes{} 
    \item[] Justification: We use both statistical significance (main paper) and confidence intervals (Appendix).

\item {\bf Experiments compute resources}
    \item[] Question: For each experiment, does the paper provide sufficient information on the computer resources (type of compute workers, memory, time of execution) needed to reproduce the experiments?
    \item[] Answer: \answerYes{} 
    \item[] Justification: See Appendix~\ref{sec:app_compute}.
    
\item {\bf Code of ethics}
    \item[] Question: Does the research conducted in the paper conform, in every respect, with the NeurIPS Code of Ethics \url{https://neurips.cc/public/EthicsGuidelines}?
    \item[] Answer: \answerYes{} 
    \item[] Justification: We have read and checked the NeurIPS Code of Ethics.

\item {\bf Broader impacts}
    \item[] Question: Does the paper discuss both potential positive societal impacts and negative societal impacts of the work performed?
    \item[] Answer: \answerYes{} 
    \item[] Justification: See Appendix~\ref{sec:app_limitations}. Also partially addressed along the main paper.
    
\item {\bf Safeguards}
    \item[] Question: Does the paper describe safeguards that have been put in place for responsible release of data or models that have a high risk for misuse (e.g., pre-trained language models, image generators, or scraped datasets)?
    \item[] Answer: \answerNA{} 
    \item[] Justification: \answerNA{}

\item {\bf Licenses for existing assets}
    \item[] Question: Are the creators or original owners of assets (e.g., code, data, models), used in the paper, properly credited and are the license and terms of use explicitly mentioned and properly respected?
    \item[] Answer: \answerYes{} 
    \item[] Justification: See Appendix~\ref{sec:app_licenses}.

\item {\bf New assets}
    \item[] Question: Are new assets introduced in the paper well documented and is the documentation provided alongside the assets?
    \item[] Answer: \answerYes{} 
    \item[] Justification: There is pseudocode in Appendix~\ref{sec:app_method_code} and the released code will provide a comprehensive README file.

\item {\bf Crowdsourcing and research with human subjects}
    \item[] Question: For crowdsourcing experiments and research with human subjects, does the paper include the full text of instructions given to participants and screenshots, if applicable, as well as details about compensation (if any)? 
    \item[] Answer: \answerNA{} 
    \item[] Justification: \answerNA{}

\item {\bf Institutional review board (IRB) approvals or equivalent for research with human subjects}
    \item[] Question: Does the paper describe potential risks incurred by study participants, whether such risks were disclosed to the subjects, and whether Institutional Review Board (IRB) approvals (or an equivalent approval/review based on the requirements of your country or institution) were obtained?
    \item[] Answer: \answerNA{} 
    \item[] Justification: \answerNA{}

\item {\bf Declaration of LLM usage}
    \item[] Question: Does the paper describe the usage of LLMs if it is an important, original, or non-standard component of the core methods in this research? Note that if the LLM is used only for writing, editing, or formatting purposes and does \emph{not} impact the core methodology, scientific rigor, or originality of the research, declaration is not required.
    \item[] Answer: \answerNA{} 
    \item[] Justification: \answerNA{}

\end{enumerate}
\else
\fi

\end{document}